\documentclass{article} 
\usepackage{preprint,times}

\usepackage{amsmath,amsfonts,bm}

\def\eqref#1{equation~\ref{#1}}

\def\1{\bm{1}}

\DeclareMathAlphabet{\mathsfit}{\encodingdefault}{\sfdefault}{m}{sl}
\SetMathAlphabet{\mathsfit}{bold}{\encodingdefault}{\sfdefault}{bx}{n}

\usepackage{hyperref}
\usepackage{url}
\usepackage{graphicx}

\usepackage{xcolor}
\usepackage{colortbl}
\usepackage{multirow}
\usepackage{multicol}

\usepackage{caption}    
\usepackage{subcaption} 
\usepackage{wrapfig}

\usepackage{enumerate}
\usepackage{hyperref}
\usepackage{color}

\hypersetup{
    colorlinks=true,
    linkcolor=blue,
}

\usepackage{color}

\definecolor{lightblue}{RGB}{150,199,237}
\definecolor{lightred}{RGB}{255,160,122}
\newcommand{\redcline}[1]{\arrayrulecolor{red}\cline{#1}\arrayrulecolor{black}}
\newcommand{\greencline}[1]{\arrayrulecolor{green}\cline{#1}\arrayrulecolor{black}}
\newcommand{\bluecline}[1]{\arrayrulecolor{blue}\cline{#1}\arrayrulecolor{black}}

\title{MME-Finance: A Multimodal Finance Benchmark for Expert-level Understanding and Reasoning}

\author{Ziliang Gan$^{1}$, Yu Lu$^{1}$, Dong Zhang$^{1}$, Haohan Li$^{1}$, Che Liu$^{2}$, Jian Liu$^{3}$, Ji Liu$^{1}$, Haipang Wu$^{1}$, \\
\AND
Chaoyou Fu$^{4}$, Zenglin Xu$^{5}$, Rongjunchen Zhang$^{1}$, Yong Dai$^{1}$ 
\\
$^{1}$HiThink Research, $^{2}$Imperial College London, $^{3}$Beihang \& $^{4}$Nanjing \& $^{5}$Fudan University}

\iclrfinalcopy
\begin{document}

\maketitle

\begin{abstract}
In recent years, multimodal benchmarks for general domains have guided the rapid development of multimodal models on general tasks. However, the financial field has its peculiarities. It features unique graphical images (e.g., candlestick charts, technical indicator charts) and possesses a wealth of specialized financial knowledge (e.g., futures, turnover rate). Therefore, benchmarks from general fields often fail to measure the performance of multimodal models in the financial domain, and thus cannot effectively guide the rapid development of large financial models. To promote the development of large financial multimodal models, we propose MME-Finance, an bilingual open-ended and practical usage-oriented Visual Question Answering (VQA) benchmark. The characteristics of our benchmark are finance and expertise, which include constructing charts that reflect the actual usage needs of users (e.g., computer screenshots and mobile photography), creating questions according to the preferences in financial domain inquiries, and annotating questions by experts with 10+ years of experience in the financial industry.
Additionally, we have developed a custom-designed financial evaluation system in which visual information is first introduced in the multi-modal evaluation process. Extensive experimental evaluations of 19 mainstream MLLMs are conducted to test their perception, reasoning, and cognition capabilities. The results indicate that models performing well on general benchmarks cannot do well on MME-Finance; for instance, the top-performing open-source and closed-source models obtain 65.69 (Qwen2VL-72B) and 63.18 (GPT-4o), respectively. Their performance is particularly poor in categories most relevant to finance, such as candlestick charts and technical indicator charts. In addition, we propose a Chinese version, which helps compare performance of MLLMs under a Chinese context. Therefore, we hope to open-source our benchmark to foster the development of multimodal models in the financial domain. The code and data will be released at \href{https://hithink-research.github.io/MME-Finance/}{https://hithink-research.github.io/MME-Finance}.

\end{abstract}

\section{Introduction}
Multimodal Large Language Models~(MLLMs)~\citep{yin2023survey}, which equip the Large Language Models~(LLMs)~\citep{radford2019language, ouyang2022training,dai2022one,touvron2023llama} with the capability of visual understanding, have experienced a revolutionary advancement recently. 
Works including Flamingo~\citep{alayrac2022flamingo}, LLaVA~\citep{liu2024visual}, CogVLM~\citep{wang2023cogvlm}, Gemini~\citep{team2023gemini}, and GPT-4o~\citep{openai2024gpt4o} have demonstrated intriguing capability to solve complex multimodal recognition and reasoning tasks. 
A reasonable and objective benchmark is of enormous significance in the success of MLLMs, which not only helps a better comparison of the performances of MLLMs but also provides valuable guidance for model optimization and real-world applications.

Early works of multimodal benchmarks, such as COCO Caption~\citep{chen2015microsoft}, GQA~\citep{hudson2019gqa}, and Flickr30k~\citep{young2014image}, have served as foundational resources for evaluating MLLMs. 
However, these benchmarks are task-specific, limiting the scope for fine-grained analysis of MLLMs' capabilities. 
More recent efforts, including MME series~\citep{mme2023,fu2024video,zhang2024mme}, MMBench~\citep{liu2023mmbench}, and MM-Vet~\citep{yu2023mm}, have shifted focus towards general multimodal tasks. 
These benchmarks comprehensively evaluate diverse capabilities of MLLMs, such as perception and reasoning, through a broader range of tasks. 
Alongside these general-purpose benchmarks, domain-specific benchmarks are rapidly emerging. 
For instance, in the medical field, benchmarks like GMAI-MMBench~\citep{chen2024gmai} and Asclepius~\citep{wang2024asclepius} have been developed, while in the autonomous driving domain, NuScenes-QA~\citep{qian2024nuscenes} and DriveLM-DATA~\citep{sima2023drivelm} are advancing research. 
These benchmarks significantly accelerated the progress of MLLMs within respective industries. 

In the financial field, understanding charts presents more unique challenges. (1) Jargon: Financial charts are filled with technical terms such as ``bullish'', ``bearish'', ``support levels'', and ``resistance levels'', which may be hard to grasp. 
(2) Complexity: Financial charts often contain a vast amount of data and information, such as the open, close, high, and low prices on a candlestick chart, along with various technical indicators and oscillators. 
(3) Diversity of Chart Types: There are multiple types of charts in the financial domain, such as line charts, bar charts, and candlestick charts, each with its specific use cases and interpretation methods. 
(4) Data Density: Financial charts may include a large number of data points, making it more difficult to identify trends and patterns. Therefore, it is challenging to comprehensively and professionally evaluate the financial capability of MLLMs. 
Benchmarks like FINANCEBENCH~\citep{islam2023financebench} and CFBenchmark~\citep{lei2023cfbenchmark} are focusing on the evaluation of LLMs. 
To the best of our knowledge, there is no multimodal benchmark in the financial area, and a significant dearth of Chinese multimodal benchmarks. Hence, a bilingual financial multimodal benchmark is urgent for promoting the development of MLLMs. 

To break this gap, we propose MME-Finance, a bilingual financial multimodal benchmark for MLLMs. 
We conduct extensive research on real-world financial application scenarios and select 6 common types of financial charts, including candlestick charts, technical indicator charts, statistical charts, tables, documents, and mixed charts. 
Based on these images and the actual usage of users in financial scenarios, we design a hierarchical series of open-ended Question Answering (QA) tasks, ranging from general visual perception like Optical Character Recognition (OCR) tasks to complex cognitive tasks such as providing investment advice.  
To ensure the quality of MME-Finance, we carefully design the annotation pipeline and invite experts with 10+ years of experience in the financial industry to conduct detailed verification of the answers. 
LLMs and MLLMs are employed for automated evaluation in MME-Finance. 
Considering the challenges of evaluating financial open-ended questions, we meticulously design the evaluation process and first introduce visual information to boost the evaluation performance. 
The effectiveness of our evaluation method has been validated through human consistency experiments.
Extensive experiments indicate existing MLLMs remain inadequate in meeting the requirements of financial tasks, where the best open-source and closed-source models have scored unsatisfactorily, with only 65.69\% (Qwen2VL-72B) and 63.18\% (GPT-4o), respectively. 
Particularly, there are three points worthy of our attention: The first point is that models encounter difficulty in some tasks, especially spatial awareness and estimated numerical calculation. The second point is that the performance related to stock charts is not good (e.g., candlestick charts and technical indicator charts), and the last is that MLLMs generally perform poorly in questions about mobile photography, which however is a relatively high-frequency use case in financial QA. We summarize our major contributions as follows:
\begin{itemize}
\item We propose MME-Finance, a novel bilingual multimodal benchmark specifically designed to evaluate the capabilities of MLLMs in the financial domain. It comprises 1,171 English and 1,103 Chinese questions, covering diverse financial image types and various multimodal capabilities, and providing a comprehensive evaluation of MLLMs' performance in the financial domain.

\item We introduce an evaluation approach of open-ended questions in the financial domain. By designing appropriate prompts for corresponding tasks and exploring evaluation methods firstly combined with image information, we propose a novel evaluation strategy that has a high consistency with humans. The strategy can serve as a reference for evaluating MLLMs for other works.

\item We conduct extensive evaluation on 19 MLLMs based on MME-Finance, revealing critical insights about the strengths and shortcomings of the current MLLMs in financial applications. The insights gained from this study provide a foundation for future research, guiding the development of more robust MLLMs capable of meeting the demands of complex financial tasks.

\end{itemize}

\section{Related Work}


\subsection{MLLMs}

Recent advancements in LLMs~\citep{radford2019language,brown2020language,ouyang2022training,touvron2023llama, chiang2023vicuna} have catalyzed significant breakthroughs in MLLMs. Utilizing pre-trained LLMs allows researchers to circumvent the resource-intensive process of training models from scratch, thereby markedly reducing computational costs. By harnessing the cognitive capabilities of LLMs, MLLMs are adept at addressing diverse multimodal challenges. To facilitate alignment between different modalities, researchers have proposed several effective connectors. Models such as BLIP-2~\citep{li2023blip}, Mini-GPT4~\citep{zhu2023minigpt}, Video-LLaMA~\citep{zhang2023video}, and X-LLM~\citep{chen2023x} employ Q-Former for the alignment of visual and textual features, while the LLaVA series~\citep{liu2024visual,liu2024improved} and VITA~\citep{fu2024vita} exploit MultiLayer Perceptrons (MLPs) for this purpose. 
Additionally, Flamingo~\citep{alayrac2022flamingo} and CogVLM~\citep{wang2023cogvlm} incorporate supplementary modules to enhance interaction and fusion between visual and textual elements. Closed-source MLLMs, such as Gemini~\citep{team2023gemini}, GPT-4V~\citep{openai2024gpt4v}, GPT-4o~\citep{openai2024gpt4o}, and Claude 3.5 Sonnet~\citep{claude2024sonnet}, demonstrate exceptional capabilities in visual understanding. 

While these MLLMs demonstrate excellent performance in standard multimodal tasks such as image captioning~\citep{vinyals2015show} and Visual Question Answering (VQA)~\citep{antol2015vqa}, their performance in specialized domains, particularly finance, remains relatively unexplored. Financial images generally present diverse content and necessitates specialized knowledge for interpretation, posing a substantial challenge for MLLMs.
  
\subsection{Multimodal Benchmarks}
MLLMs have demonstrated exceptional capabilities across various complex tasks. Objective and accurate quantification of these capabilities is essential for informing future development trajectories, making the establishment of comprehensive benchmarks significant for advancing MLLMs. Traditional multimodal benchmarks typically focus on single tasks, for instance, COCO Caption~\citep{chen2015microsoft} and Flickr30k~\citep{young2014image} address captioning, while GQA~\citep{hudson2019gqa}, VQAv2~\citep{goyal2017making}, and VizWiz~\citep{gurari2018vizwiz} pertain to VQA. Other benchmarks assess specific capabilities, such as TextCaps~\citep{sidorov2020textcaps} and Tap~\citep{yang2021tap} for scene text understanding, and VCR~\citep{zellers2019recognition} for commonsense reasoning. Subsequent benchmarks have expanded in both data volume and task categories. The MME benchmark~\citep{mme2023} proposes a comprehensive assessment across 14 perception and cognition tasks, while MMBench~\citep{liu2023mmbench} constructs over 3,000 multiple-choice image question pairs encompassing 20 abilities. SEED-Bench~\citep{li2023seed} and SEED-Bench-2~\citep{li2023seed2} further scale the sample sizes to 19,000 and 24,371 QA pairs from diverse scenarios, respectively. Collectively, these benchmarks provide thorough evaluations of MLLMs' capacities to tackle general multimodal challenges. 

However, the performance evaluation of MLLMs in specific domains remains underexplored, particularly in finance. Existing benchmarks like FINANCEBENCH~\citep{islam2023financebench} and CFBenchmark primarily~\citep{lei2023cfbenchmark} assess LLMs rather than MLLMs.

\section{MME-Finance}

In this section, we introduce MME-Finance by first elaborating on the design philosophy of the benchmark in Section~\ref{subsec:level}, followed by a detailed description of the data collection in Section~\ref{subsec:collection}, data annotation in Section~\ref{subsec:annotation}, and the statistics in Section~\ref{subsec:statistics}. 
Finally, we expound on the evaluation method of MME-Finance in Section~\ref{subsec:evaluation}.

\subsection{Hierarchical Ability Levels of MME-Finance}
\label{subsec:level}

The abilities of MLLMs can be divided into three categories: visual understanding, logical reasoning, and complex cognition. 
MME-Finance references these categories and organizes a three-tier ability structure. 
Specifically, we define perceptual ability as the low-level capacity for extracting and interpreting visual information from images. 
This foundational ability supports other advanced capabilities. 
To evaluate the perceptual ability, MME-Finance employs tasks such as image captioning, OCR, entity recognition, and spatial awareness.
As a middle-level ability, reasoning encompasses financial-related numerical reasoning.
MME-Finance evaluates this ability through tasks involving both estimated and accurate numerical calculations.
Cognition is considered as a high-level ability, which requires integrating perceptual and reasoning skills with domain-specific financial knowledge to generate reasonable answers. 
The corresponding tasks, typically complex and requiring expert-level financial insight, include reason explanation, risk warning, investment advice, and financial knowledge QA. It should be noted that some cognitive tasks are insufficient to answer based solely on image information. For such tasks, MME-Finance provides additional background information retrieved via web searches, supplementing the images and questions. 
These tasks require MLLMs to synthesize both image content and background information to derive the correct answers.
Additionally, to assess the capability to handle hallucinations in MLLMs, MME-Finance includes the not applicable task, which means the answer is not applicable for the question.

\begin{figure}[h]
\begin{center}
\includegraphics[width=1.0\columnwidth]{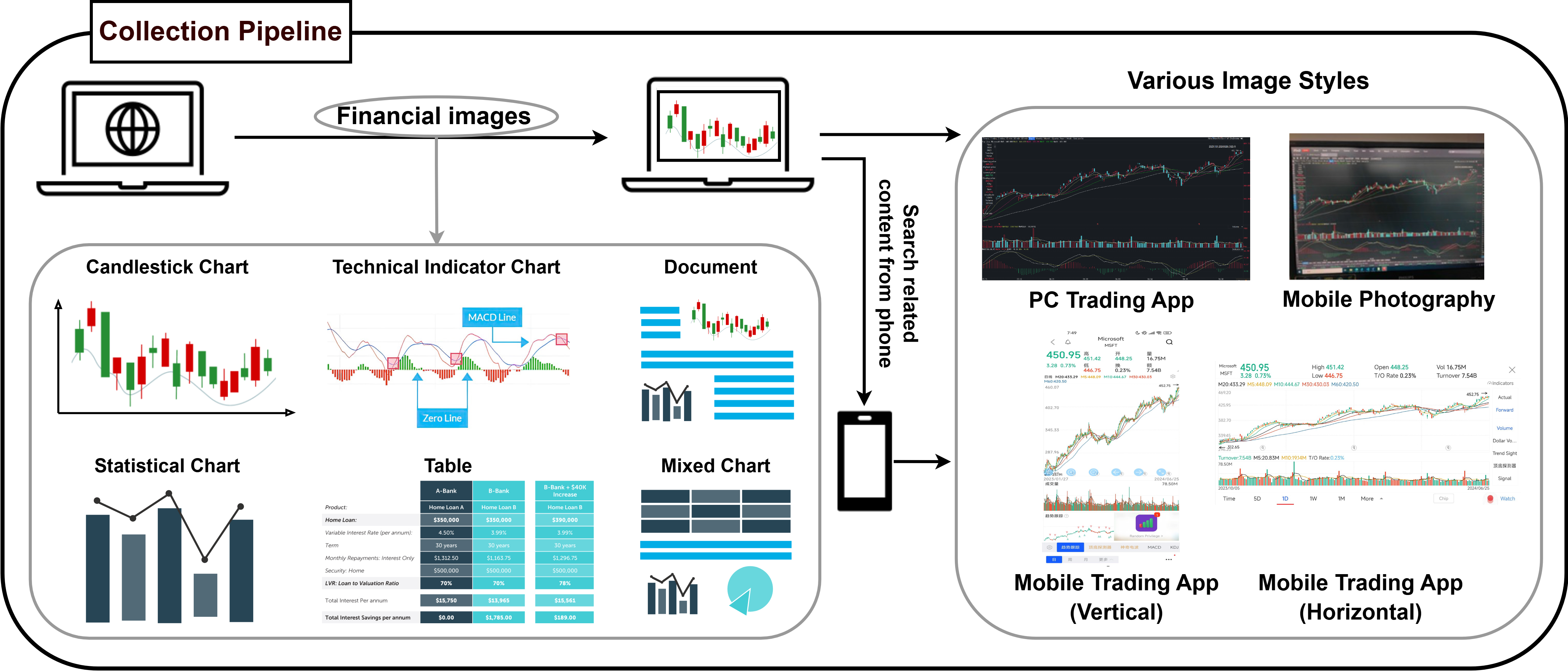}
\end{center}
\caption{Data collection pipeline of MME-Finance.}
\label{fig:fig1}
\end{figure}

\subsection{Data Collection}
\label{subsec:collection}

In MME-Finance, we collect financial images from various mainstream platforms.
Figure~\ref{fig:fig1} illustrates the data collection pipeline. 
First, we identify relevant financial pages on a computer and use screenshot tools to capture the appropriate areas. Then, we use mobile devices to photograph the corresponding sections. 
Next, we search for the same content on mobile applications and capture screenshots using smartphones. 
The inclusion of diverse image styles, including computer screenshots, mobile photographs, and vertical and horizontal mobile screenshots, is intended to simulate real-world application scenarios. 
MME-Finance categorizes the collected images into six types: candlestick charts, technical indicator charts, statistical charts, tables, documents, and mixed charts, where a mixed chart includes at least two of other types. 
These images cover a broad spectrum of financial scenarios, enabling MME-Finance to evaluate MLLMs' ability to address challenges in this domain comprehensively.

\begin{figure}[h]
\begin{center}
\includegraphics[width=1.0\columnwidth]{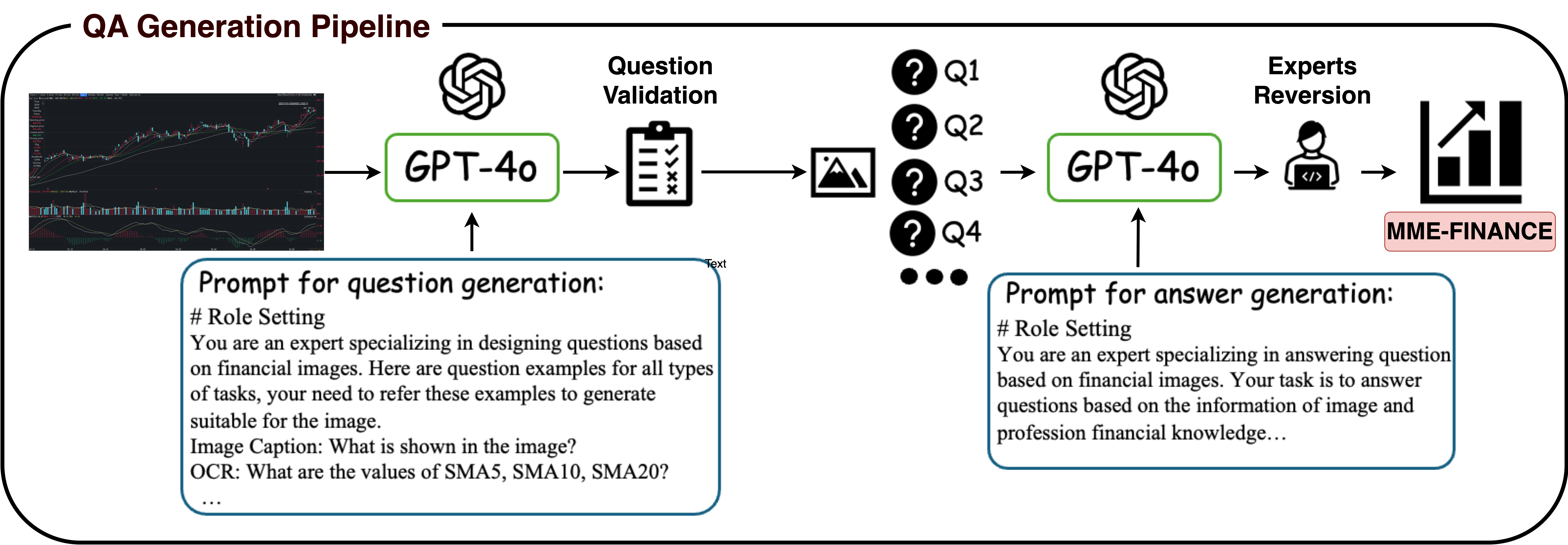}
\end{center}
\caption{Annotation generation pipeline of MME-Finance.}
\label{fig:fig2}
\end{figure}

\subsection{QA Generation}
\label{subsec:annotation}

To generate high-quality QA pairs for MME-Finance, each QA pair underwent at least two stages of manual evaluation. 
Figure~\ref{fig:fig2} illustrates the QA generation pipeline. We first design several question examples for each task. 
Then we utilize GPT-4o to generate candidate questions for every image based on the example questions. 
We meticulously review the questions and correct inappropriate ones. 
In the answer generation stage, we also use GPT-4o to generate preliminary answers based on questions and images. 
We check all the answers manually and correct the wrong ones. 
The complex subjective questions are evaluated by a panel of three finance researchers, each with over 10 years of experience.
The reference answer is confirmed when the reviewers reach a consensus. 
After this process, financial experts conduct an in-depth examination and refinement. The quality of MME-Finance is significantly enhanced through the manual review mechanism. 

\begin{wraptable}{r}{0.5\textwidth}
\caption{Statistic of the number of samples in different capabilities and tasks.}
\label{tab:stat}
\centering
\setlength{\tabcolsep}{2.0pt}
\begin{tabular}{lc}
\hline
 \textbf{Statistic} & \textbf{Number}
\\ \hline 
\textbf{Perception} &  734  \\
~- Image Caption &164  \\
~- OCR &178  \\
~- Entity Recognition &163 \\
~- Spatial Awareness  &229\\ \hline
\textbf{Reasoning} & 175 \\
~- Accurate Numerical Calculation & 133 \\
~- Estimated Numerical Calculation & 42 \\ \hline
\textbf{Cognition} & 240 \\ 
~- Risk Warning & 22 \\
~- Investment Advice & 53 \\
~- Reason Explanation & 18 \\
~- Financial Question Answer & 147 \\ \hline
\textbf{Hallucination } & 22 \\
~- Not Applicable &22 \\
\hline
\end{tabular}
\end{wraptable}

\subsection{Statistics}
\label{subsec:statistics}

\begin{figure}[t]
    \centering
    \begin{subfigure}[b]{0.48\textwidth} 
        \centering
        \includegraphics[width=\textwidth]{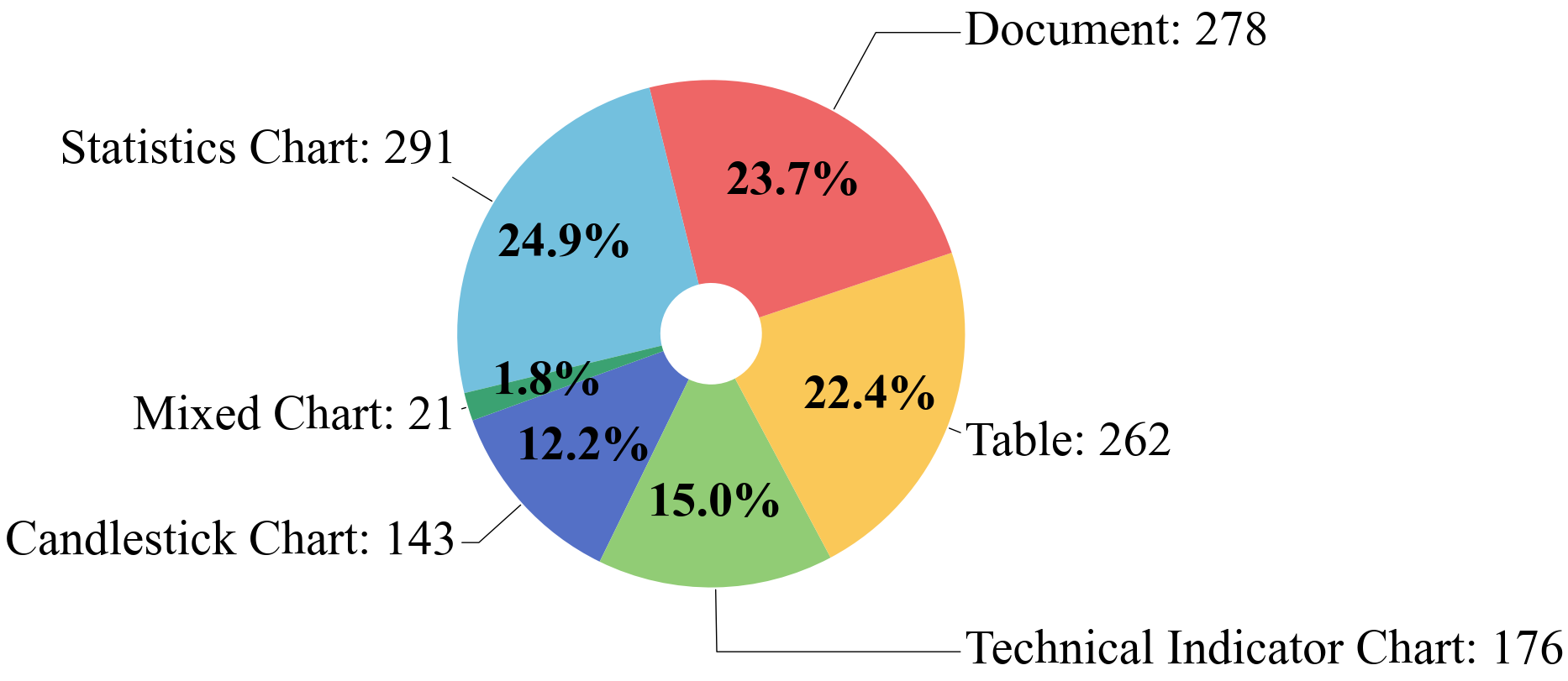} 
        \caption{}
        \label{fig:subfig3a}
    \end{subfigure}
    \hspace{0.05\textwidth}
    \begin{subfigure}[b]{0.45\textwidth} 
        \centering
        \includegraphics[width=\textwidth]{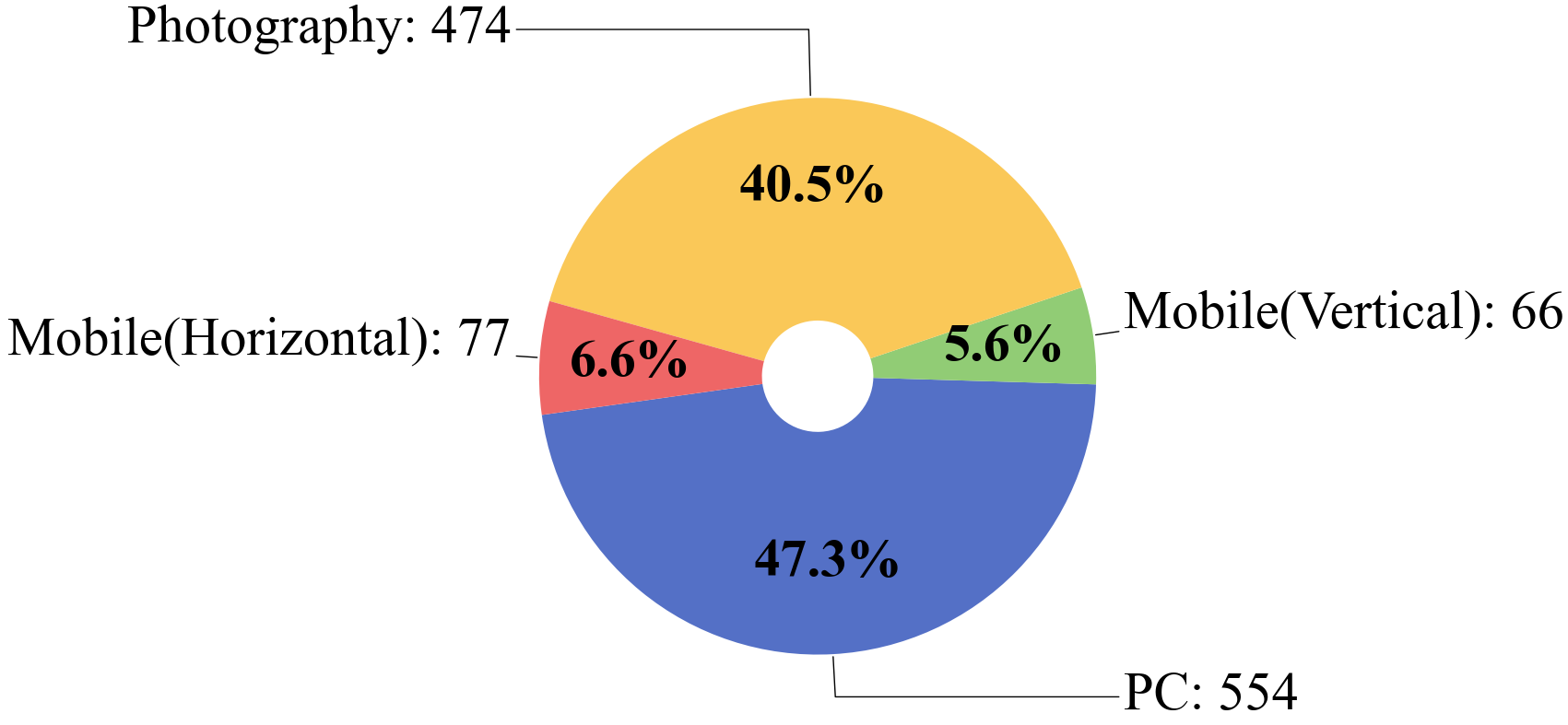} 
        \caption{}
        \label{fig:subfig3b}
    \end{subfigure}
    \caption{Distribution of different (a) types and (b) styles of images.}
    \label{fig:mainfig}
    \vspace{-3mm}
\end{figure}

As shown in Table~\ref{tab:stat}, English MME-Finance contains 1,171 image-question-answer pairs spanning 11 distinct tasks, categorized into 3 ability levels as detailed in Section~\ref{subsec:level}. 
In addition, MME-Finance incorporates questions aimed at evaluating hallucinations~\citep{yin2023woodpecker} of MLLMs. 
The number of samples per task varies from 18 to 229, with the ``Spatial Awareness" task containing the most and ``Reason Explanation" the fewest. 
Figure~\ref{fig:mainfig}(\subref{fig:subfig3a}) illustrates the distribution of the 6 image types, where statistical charts account for the main proportion, while mixed charts are the least. 
Figure~\ref{fig:mainfig}(\subref{fig:subfig3b}) displays the distribution of 4 image styles. 
Computer screenshots and mobile photographs constitute similar proportions, representing 47.3\% and 40.5\% of the total, respectively. 
Vertical and horizontal mobile screenshots contain approximately sample sizes. 
The statistic results of Chinese version are shown in the appendix.

\subsection{Evaluation Method}
\label{subsec:evaluation}

\begin{figure}[h]
\begin{center}

\includegraphics[width=1.0\columnwidth]{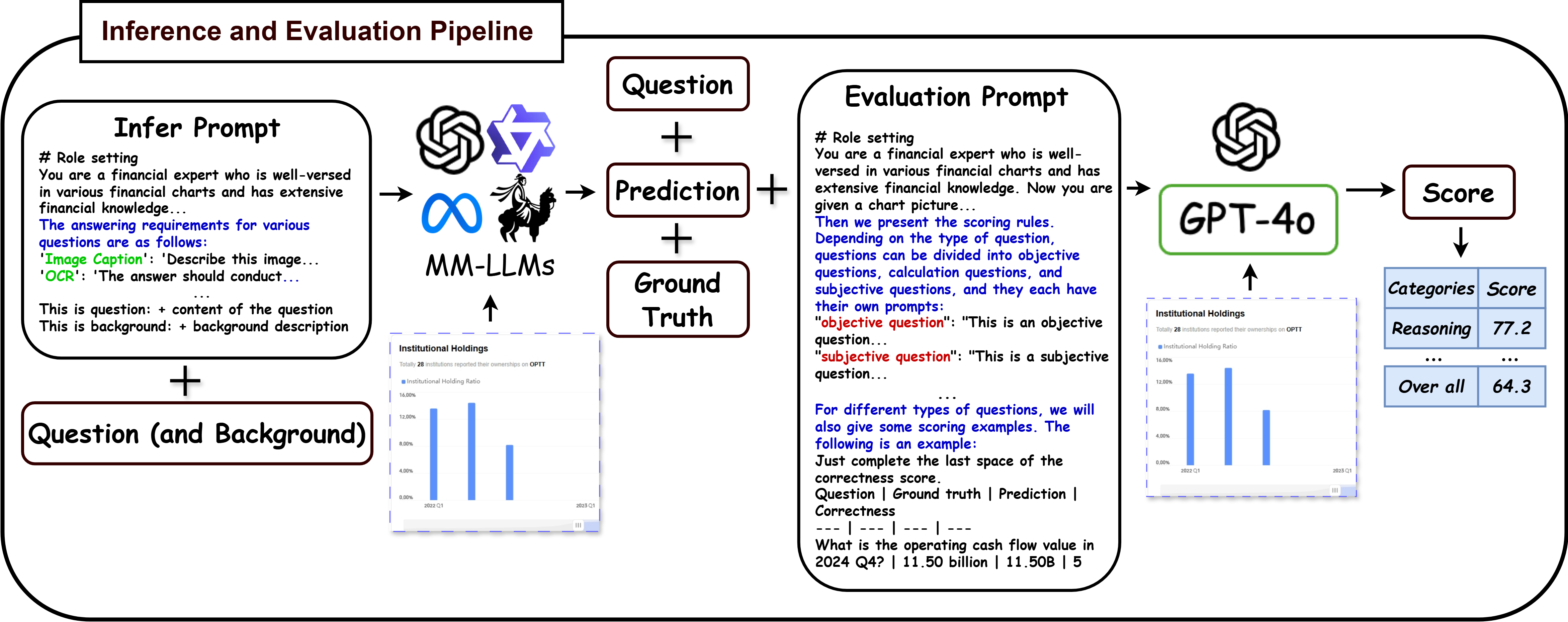}
\end{center}
\caption{Inference and evaluation pipeline of MME-Finance. We first input the image and question prompt into the MLLMs. Then we feed the image and evaluation prompt into GPT-4o to obtain scores. The question and evaluation prompts are all designed individually for each task category.}
\label{fig:fig4}
\end{figure}

MME-Finance's QA format is intentionally open-ended to reflect the complexity of real-world financial scenarios. However, evaluating open-ended responses presents greater challenges compared to multiple-choice questions. 
To accurately evaluate the capabilities of MLLMs, we design a comprehensive evaluation process tailored to the characteristics of our benchmark. 
As shown in Figure~\ref{fig:fig4}, during the inference phase, prompts are crafted to constrain the output formats of MLLMs, thereby facilitating a more standardized evaluation. 
Drawing inspiration from the evaluation methodology used in MM-Vet~\cite{yu2023mm}, we employ an LLM-based evaluation system to compare model predictions with the ground truth and to assign a score. 
The scoring system is divided into six levels, ranging from 0 (completely incorrect) to 5 (fully correct), with the overall score being the average across all samples. 
Given the diversity in response formats across different tasks, we develop task-specific evaluation prompts to ensure accurate assessments. 
Additionally, a few-shot approach is employed to define scoring metrics using in-context examples, which aids the model in producing more accurate evaluation scores. Our experimental results demonstrate that the LLM-based evaluator, particularly GPT-4o, achieves the highest consistency with human evaluators.

\section{Experiment}
In this section, we introduce the experimental setup for evaluating MLLMs firstly in Section~\ref{subsec:Setup}, followed by an exhibition and analysis of the experimental results. The main result analysis is presented in Section~\ref{subsec:MainR}, followed by a detailed analysis of the ability dimension in Section~\ref{subsubsec:AbiDim} and image type and style dimension in Section~\ref{subsubsec:ImagDim}. Finally, Section~\ref{subsec:EvaAna} elaborates on our analysis of LLM as an evaluator.

\subsection{Experimental Setup}\label{subsec:Setup}

We utilize MME-Finance to evaluate two types of MLLMs, (1)~Open-Source MLLMs including CogVLM2~\citep{hong2024cogvlm2}, Qwen2-VL~\citep{Qwen2-VL}, MiniCPM-V 2.6~\citep{yao2024minicpm}, Phi3-Vision~\citep{abdin2024phi}, Phi3.5-Vision~\citep{abdin2024phi}, LLaMA3.2~\citep{meta2024llama3.2}, LLaVA-NEXT~\citep{liu2024llavanext}, YiVL~\citep{ai2024yi}, and InternVL2~\citep{chen2024far}; (2)~Proprietary MLLMs including GPT-4o, GPT-4o mini~\citep{openai2024gpt4o}. 
The inference prompts are the same for all MLLMs for a fair comparison, and a zero-shot setting is adopted. 
We fill the prompt template with image, question, ground truth, and response from an MLLM, and take the filled prompt into an LLM-based evaluator for generating a score range from 0 to 5 for one sample. 
The scores are multiplied by 20\% to be normalized.  

\subsection{Main Results}\label{subsec:MainR} 

Table~\ref{tab:tab2} shows the results of various MLLMs on English MME-Finance from the view of each task. Performance across the MLLMs varies significantly, with many models exhibiting low accuracy, highlighting the challenging nature of the MME-Finance benchmark. Among the evaluated models, Qwen2VL-72B achieves the best overall performance with 65.69\% accuracy, excelling in most tasks, particularly OCR and ANC. Proprietary MLLM, i.e., GPT-4o, ranks second overall but surpasses Qwen2VL-72B in all cognition-related tasks. This suggests that GPT-4o’s superior language processing capabilities give it an advantage in tasks requiring complex reasoning. Additionally, our findings support the observation from MMBench~\citep{liu2023mmbench} that \emph{the size of the language model has a significant impact on performance}. For instance, larger models in the same series, such as LLaVA-NEXT-13B compared to LLaVA-NEXT-7B, consistently demonstrate better results.

\begin{table}[t]
\caption{Evaluation results on English MME-Finance for all tasks. Abbreviations adopted: IC for Image Caption; ER for Entity Recognition; SA for Spatial Awareness; FQA for Financial Question Answer; ANC for Accurate Numerical Calculation; ENC for Estimated Numerical Calculation; RW for Risking Warning; IA for Investment Advice; RE for Reason Explaination; NA for Not Applicable. 
The first, the second, and the third highest values are highlighted by \colorbox{orange!40}{orange}, \colorbox{cyan!40}{blue}, and \colorbox{green!40}{green} backgrounds. All numbers are denoted in \% with the max value of 100\%.}
\label{tab:tab2}
\begin{center}
\renewcommand{\arraystretch}{1.2}
\setlength{\tabcolsep}{3.0pt}
\small
\begin{tabular}{ccccccccccccc}
\hline
\multirow{2}{*}{\textbf{Model}} & \multirow{2}{*}{\textbf{Overall}} & \multicolumn{4}{c}{\textbf{Perception}}    & \multicolumn{2}{c}{\textbf{Reasoning}}  & \multicolumn{4}{c}{\textbf{Cognition}} & \multirow{2}{*}{\textbf{NA}}\\ 
\redcline{3-6} \bluecline{7-8} \greencline{9-12}
& &\textbf{IC} & \textbf{OCR} & \textbf{ER} & \textbf{SA}  & \textbf{ANC} & \textbf{ENC} & \textbf{RW} & \textbf{IA} & \textbf{RE} & \textbf{FQA}
\\ \hline 
\multicolumn{13}{c}{Open source MLLMs} \\
Yi-VL-34B &17.57  &29.39  &1.46  &3.93  &8.73   &5.56  &11.43  &42.73  &35.09 &58.89 &47.48 &36.36\\
CogVLM2-19B &46.32  &67.32  &61.24  &35.83  &16.59    &44.51  &33.33  &59.09  &52.83 &31.11 &58.64 &\cellcolor{cyan!40}93.64\\
InternVL2-2B &37.42  &59.63  &46.97  &21.23  &18.52    &28.27  &19.05  &59.09  &50.94  &60.00 &51.70 &33.63 \\
InternVL2-4B &47.69  &67.44  &58.88  &33.74  &18.95    &55.49  &30.48  &68.18  &54.34  &64.44 &60.95 &59.09 \\
InternVL2-8B &53.58  &71.71  &68.43  &38.28 &25.33 &62.86    &37.14  &72.73  &60.75  &\cellcolor{green!40}76.67 &63.13 &61.82 \\
InternVL2-76B &\cellcolor{green!40}61.62  &\cellcolor{cyan!40}83.17  &\cellcolor{green!40}77.64  &\cellcolor{green!40}47.60 &\cellcolor{orange!40}30.31 &\cellcolor{green!40}70.08    &\cellcolor{green!40}41.90  &75.45  &\cellcolor{cyan!40}66.42  &\cellcolor{green!40}76.67 &72.24 &79.09 \\
LLaMA3.2-11B &42.51  &62.44  &39.10  &32.02  &14.50    &55.79  &37.14  &60.00  &50.57  &68.89 &57.55 &61.82 \\
LLaMA3.2-90B &48.76  &64.27  &46.74  &41.27  &25.85  &55.64  &22.86  &63.64  &61.13  &64.44 &65.58 &81.82 \\
LLaVA-NEXT-7B &28.18  &58.41  &22.81  &14.85  &11.09  &7.07  &10.00 &45.45  &47.55  &12.22  &54.97  &55.45 \\
LLaVA-NEXT-13B &31.37  &62.68  &25.39  &22.58  &10.31  &12.63  &9.05 &47.27  &40.00  &12.22  &59.46  &78.18 \\
MiniCPM2.6 &51.65  &71.22  &63.71  &37.67  &24.37   &55.64  &21.43  &72.73  &58.87 &66.67 &66.80 &77.27\\
Phi3-Vision &46.69  &69.88  &57.64  &28.34  &18.08   &47.52  &34.76  &65.45  &58.11 &68.89 &57.41 &\cellcolor{orange!40}100.0\\
Phi3.5-Vision &38.99  &67.56  &33.03  &18.90  &20.52   &32.33  &19.52  &67.27  &55.85 &72.22 &54.42 &\cellcolor{cyan!40}93.64\\
Qwen2VL-2B &44.42  &62.07  &66.07  &28.47  &20.09   &44.36  &23.33  &53.63  &44.53  &58.89  &53.47 &68.18 \\
Qwen2VL-7B &44.44  &62.19  &64.49  &26.50  &19.04   &45.56  &27.62  &57.27  &48.30  &58.89 &54.97 &68.18 \\
Qwen2VL-72B &\cellcolor{orange!40} 65.69  &\cellcolor{green!40}82.56  &\cellcolor{orange!40}87.52  &\cellcolor{orange!40}55.46  &\cellcolor{cyan!40}27.16   &\cellcolor{orange!40}83.76  &40.95  &\cellcolor{cyan!40}78.18  &\cellcolor{green!40}65.66  &\cellcolor{cyan!40}77.78  &\cellcolor{cyan!40}75.37 &\cellcolor{green!40}90.91 \\ 
 \hline
\multicolumn{13}{c}{Proprietary MLLMs} \\
GPT-4o-05-13 &42.85  &71.34  &28.09  &28.22  &19.65  &31.73  &36.19 &\cellcolor{green!40}76.36  &62.26  &75.56  &71.43  &81.82 \\
GPT-4o-mini &57.34  &79.15  &68.99  &40.25  &24.72  &63.31  &\cellcolor{cyan!40}43.81 &73.64  &64.53  &\cellcolor{cyan!40}77.78  &\cellcolor{green!40}73.20  &\cellcolor{orange!40}100.0 \\
GPT-4o &\cellcolor{cyan!40}63.18  &\cellcolor{orange!40}83.66  &\cellcolor{cyan!40}79.21  &\cellcolor{cyan!40}49.81  &\cellcolor{green!40}27.07  &\cellcolor{cyan!40}71.88  &\cellcolor{orange!40}44.76  &\cellcolor{orange!40}84.54  &\cellcolor{orange!40}70.57  &\cellcolor{orange!40}80.00  &\cellcolor{orange!40}76.87  &\cellcolor{cyan!40}93.64 \\ \hline

\end{tabular}
\end{center}
\end{table}

\subsubsection{Ability Dimensional Analysis}
\label{subsubsec:AbiDim}

\begin{figure}[h]
\begin{center}
\includegraphics[width=1.0\columnwidth]{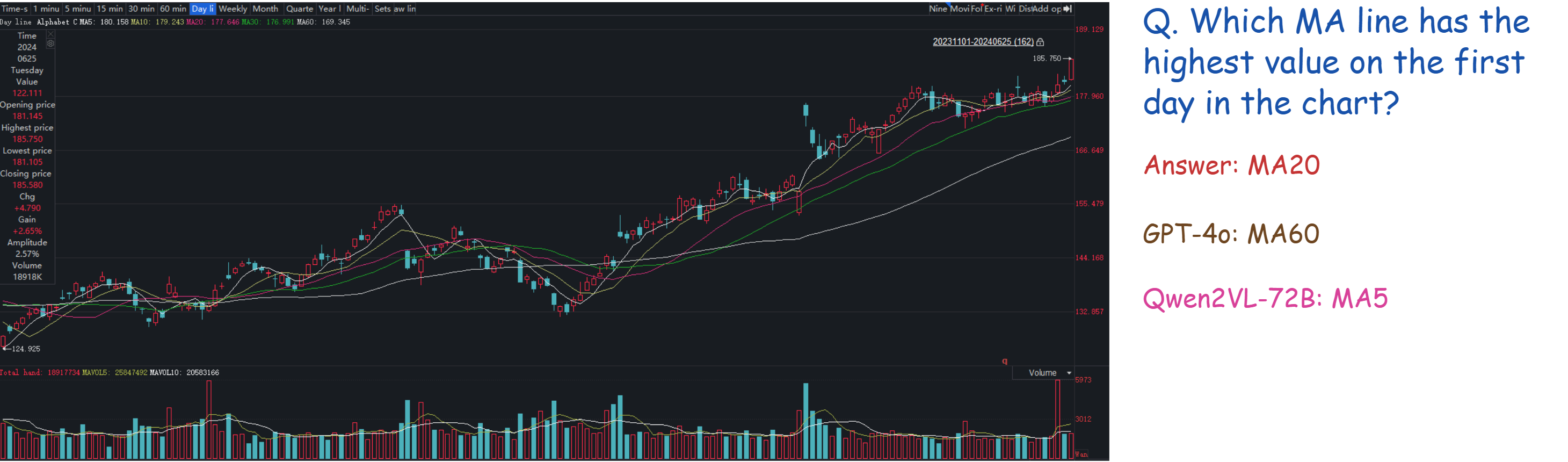}
\end{center}
\caption{Hard example in SA task of perception capability.}
\label{fig: SA}
\end{figure}

\noindent\textbf{Perception.} The ``Perception'' ability encompasses four tasks: Image Captioning (IC), Optical Character Recognition (OCR), Entity Recognition (ER), and Spatial Awareness (SA), all of which primarily focus on visual understanding. MLLMs tend to perform relatively well in IC and OCR tasks, suggesting that current models exhibit satisfactory general visual perception capabilities. 
However, the SA task proves to be the most challenging, with an highest accuracy of only 30.31\%.
This difficulty likely stems from the need for fine-grained perceptual abilities in the SA task. For example, as shown in Figure~\ref{fig: SA}, the task requires identifying the highest Moving Average (MA) line. Several MA lines are closely positioned in the image, making it difficult for the models to distinguish between them. 
\emph{This suggests that although MLLMs exhibit competence in general visual tasks, there remains substantial room for improvement in more fine-grained visual perception.}

\begin{figure}[h]
\begin{center}
\includegraphics[width=1.0\columnwidth]{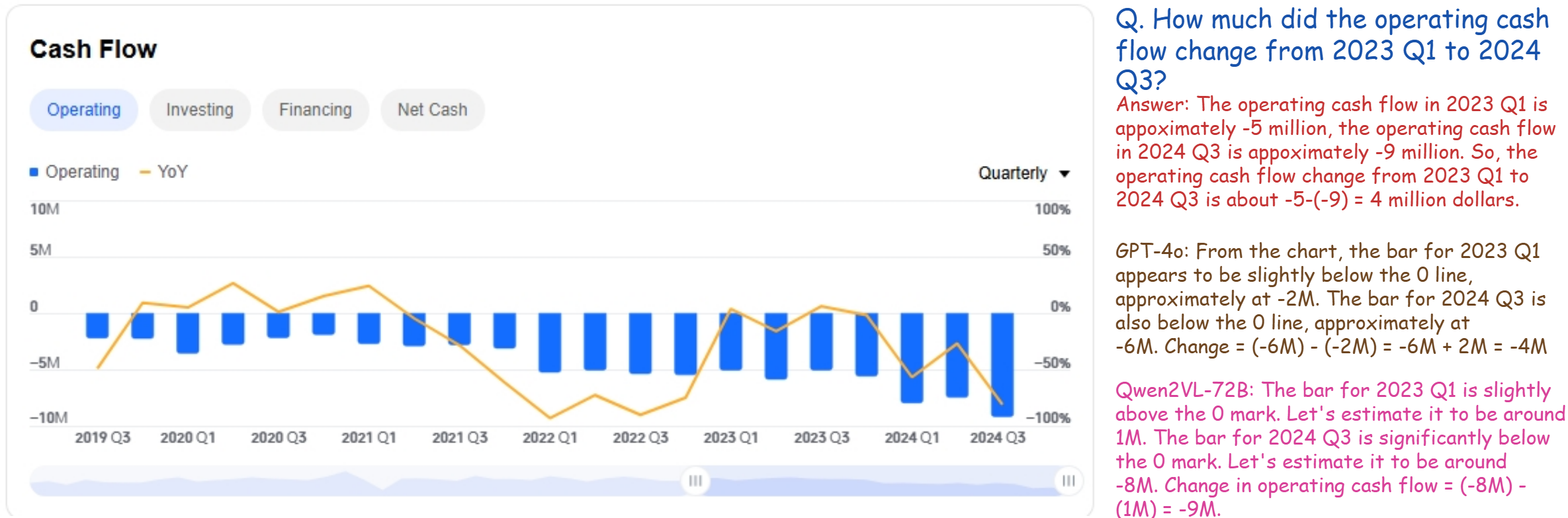}
\end{center}
\caption{Hard example in ENC task of reasoning capability.}
\label{fig:fig6}
\end{figure}

\noindent\textbf{Reasoning.} 
The ``Reasoning'' ability consists of two tasks: Accurate Numerical Calculation (ANC) and Estimated Numerical Calculation (ENC), both of which focus on mathematical and logical reasoning. Among these, the ENC task is significantly more challenging. This difficulty arises from the inherent complexity of estimating reasonable numerical values, a ability that current MLLMs struggle with. 
As shown in Table~\ref{tab:tab2}, the best-performing model, i.e., Qwen2VL-72B, achieves only 40.95\% accuracy on the ENC task, which is much lower compared to 83.76\% on the ANC task. This discrepancy highlights the continued difficulty MLLMs face in handling estimation-based reasoning problems.
As depicted in Figure~\ref{fig:fig6}, the exact numerical values are not explicitly presented in the image, requiring the model to infer these values based on contextual clues, such as spatial relationships. 
\emph{Hence, the inability to reasonably estimate such values also remains a critical limitation of current MLLMs.}

\begin{table}[t]
\caption{Evaluation results on English MME-Finance for different types and styles of images. Abbreviations adopted: Candle. for Candlestick chart; Tech. for Technical indicator chart; Stat. for Statistical chart; Tab. for Table; Doc. for Document; Mixed for Mixed chart; CS for Computer Screenshot; MP for Mobile Photograph; VS for Vertical Screenshot on Mobile; HS for Horizontal Screenshot on Mobile. The first, the second, and the third highest values are highlighted by \colorbox{orange!40}{orange}, \colorbox{cyan!40}{blue}, and \colorbox{green!40}{green} backgrounds. All numbers are denoted in \% with the max value of 100\%.}
\label{tab:tab3}
\begin{center}
\renewcommand{\arraystretch}{1.2}
\setlength{\tabcolsep}{5pt}
\small
\begin{tabular}{lllllll|llll}
\hline
\textbf{Model} &\textbf{Candle.} & \textbf{Tech.} & \textbf{Stat.} & \textbf{Tab.} & \textbf{Doc.} & \textbf{Mixed} & \textbf{CS} & \textbf{MP} & \textbf{VS} & \textbf{HS} 
\\ \hline 
\multicolumn{11}{c}{Open source MLLMs} \\
Yi-VL-34B &23.64  &16.36  &18.76 &15.42  & 14.89  &32.38  &19.42  &14.39  &26.06 &16.62 \\
CogVLM2-19B &39.44  &35.57  &52.30  &50.38  &45.76    &57.14  &47.33  &44.22  &49.70 &49.09\\
InternVL2-2B &30.35  &33.18  &38.62  &40.00  &38.49    &58.10  &40.36  &34.73  &35.45  &34.55 \\
InternVL2-4B &35.38  &38.98  &51.48  &54.66  &47.77    &63.81  &50.87  &44.85  &43.64  &45.71 \\
InternVL2-8B &42.38  &45.00  &60.41  &57.79 &52.59 &67.62    &56.39  &51.56  &48.79  &49.87 \\
InternVL2-76B &\cellcolor{green!40}55.52  &47.50  &63.02  &\cellcolor{cyan!40}70.84 &\cellcolor{cyan!40}63.09 &67.62    &\cellcolor{green!40}62.78  &\cellcolor{cyan!40}61.73  &54.54  &58.70 \\
LLaMA3.2-11B &35.24  &31.59  &47.63  &50.92  &39.42   &48.57  &45.16  &39.07  &38.79  &47.79\\
LLaMA3.2-90B &40.56  &40.11  &51.20  &58.17  &45.83  &64.76  &50.14  &46.33  &46.06  &56.10 \\
LLaVA-NEXT-7B &29.65  &23.52  &28.80  &28.32  &28.34  &44.76  &28.45 &26.08  &32.73  &35.32 \\
LLaVA-NEXT-13B &27.27  &26.36  &33.68  &32.14  &32.95  &39.05  &32.67 &29.20  &30.91  &35.84\\
MiniCPM2.6 &45.03  &45.00  &54.23  &58.63  &49.42   &59.05  &52.09  &50.51  &45.45 &\cellcolor{green!40}60.78\\
Phi3-Vision &37.62  &40.00  &49.48  &49.54  &48.71   &62.86  &49.75  &43.08  &40.30 &52.21\\
Phi3.5-Vision &32.73  &30.45  &46.25  &38.24  &39.21   &59.05  &44.73  &32.28  &41.52 &36.88\\
Qwen2VL-2B &38.74  &40.80  &46.60  &46.26  &44.68   &57.14  &45.13  &43.71  &38.79  &48.57  \\
Qwen2VL-7B &39.72  &41.70  &46.60  &46.11  &44.03   &54.29  &44.73  &44.09  &36.97  &50.91 \\
Qwen2VL-72B &\cellcolor{orange!40}60.12  &\cellcolor{orange!40}60.11  &\cellcolor{cyan!40}65.15 &\cellcolor{orange!40}71.73  &\cellcolor{orange!40}66.04  &\cellcolor{orange!40}74.24   &\cellcolor{orange!40}67.65  &\cellcolor{orange!40}62.78  &\cellcolor{cyan!40}68.48  &\cellcolor{cyan!40}67.01  \\ 
 \hline
\multicolumn{11}{c}{Proprietary MLLMs} \\
GPT-4o-05-13 &44.62  &32.84  &52.99  &37.18 &41.65  &60.95  &46.43  &37.26  &47.27  &47.79  \\
GPT-4o-mini &51.89  &\cellcolor{green!40}50.91  &\cellcolor{green!40}63.37  &60.46 &54.10  &\cellcolor{green!40}68.57  &59.71  &53.45  &\cellcolor{green!40}62.12  &60.00  \\
GPT-4o &\cellcolor{cyan!40}58.32  &\cellcolor{cyan!40}55.68  &\cellcolor{orange!40}67.84  &\cellcolor{green!40}68.55  &\cellcolor{green!40}59.71  &\cellcolor{cyan!40}73.33  &\cellcolor{cyan!40}65.67  &\cellcolor{green!40}58.31  &\cellcolor{orange!40}69.09  &\cellcolor{orange!40}70.13  \\ \hline
\end{tabular}
\end{center}
\end{table}

\noindent\textbf{Cognition.} The ``Cognition" task, consisting of Risk Warning (RW), Investment Advice (IA), Reason Explanation (RE), and Financial Knowledge QA (FQA), assesses the ability of MLLMs to make complex financial decisions. Due to the inherently subjective nature of these questions, the performance variance among different models is smaller compared to other tasks. This suggests that current MLLMs demonstrate a basic competence in financial reasoning. GPT-4o achieves the highest overall score across those 4 tasks, indicating its superior capability in handling financially complex and subjective decision-making scenarios. 

\noindent\textbf{Hallucination Problem.} 
The Not Applicable (NA) task is specifically designed to assess the hallucination of MLLMs. For this task, we develop an inference prompt that explicitly informs the models that they can respond with ``Not Applicable'' if they determine that no suitable answer is attainable. As shown in Table~\ref{tab:tab2}, models such as CogVLM2-19B, Phi3.5-Vision, Qwen2VL-72B, and GPT-4o demonstrate a strong ability to discern whether a question is answerable, thereby mitigating the hallucination issue. In contrast, models like Yi-VL-34B, InternVL2-2B, and LLaVA-NEXT-7B exhibit severe hallucination problems. Since the prompt reminds MLLMs of the ``Not Applicable" option in the NA task, the task difficulty is somewhat reduced. To explore hallucination issues more comprehensively, we modified the prompt to allow ``Not Applicable" response across all types of tasks. This led to a rise in false negatives in most MLLMs, where models incorrectly mark questions with valid answers as unanswerable, suggesting that hallucination remains a significant challenge for many MLLMs. The corresponding experimental results are detailed in the appendix.

\subsubsection{Image Type and Style Dimensional Analysis}
\label{subsubsec:ImagDim}
Table~\ref{tab:tab3} presents the performance of various MLLMs from the perspective of image types and styles. Notably, most models exhibit poor performance on candlestick charts and technical indicator charts. This can be attributed to the specialized nature of these charts, which demand domain-specific knowledge that current MLLMs struggle to interpret. 
Regarding image styles, MLLMs exhibit suboptimal performance when applied to mobile photographs, primarily attributed to the lower resolution of images captured by phones, which hampers the visibility of crucial details.
Furthermore, many of these photos are taken at oblique angles, leading to incomplete or extraneous visual information. 
\emph{Given the prevalence of such image in real-world applications, it is imperative to enhance the proficiency of MLLMs in order to effectively process and analyze mobile photographs.}

\subsection{Analysis of Evaluators}
\label{subsec:EvaAna}

For a fair comparison of the tested MLLMs, we conduct extensive experiments to assess the effectiveness of various evaluators. 
First, we select 100 samples and generate corresponding outputs from MiniCPM2.6. To ensure objectivity, each sample is scored by three experienced experts. The final score for each sample is determined by mode (e.g., 3 for scores of 2, 3, and 3) or mean (e.g., 2 for scores of 1, 2, and 3). If the score variance exceeds 2, the sample will undergo further review to determine a final score. Then, Spearman’s rank correlation coefficient and average absolute differences are calculated to indicate the evaluators’ performance.
As shown in Table~\ref{evaluator}, GPT-4o with image input achieves the highest Spearman’s rank correlation coefficient and the lowest average absolute difference, indicating superior performance. 
GPT-4-Turbo also demonstrates strong performance, significantly outperforming GPT-3.5-Turbo and o1-preview. 
Among the open-source evaluators, Qwen2VL-72B demonstrates performance comparable to GPT-4o, substantially surpassing CogVLM2 and MiniCPM2.6. \emph{Therefore, Qwen2VL-72B can serve as a cost-effective alternative to GPT-4o.}
Furthermore, we observe that most evaluators perform better with additional image inputs. We argue that this improvement is due to the images provide evaluators with additional information. We further divide the questions into two categories: subjective and objective. From Table~\ref{sub_obj_evaluator}, it can be seen that the GPT-4o evaluator exhibits higher consistency in scoring objective questions, and the inclusion of images notably improves accuracy in evaluating subjective questions. 

\begin{table}[t]
\caption{The comparison of Spearman's rank correlation coefficient~(Sp.) and average absolute differences~($\Delta$) between the evaluation scores of various evaluators and human-annotated scores. Larger Sp. and smaller $\Delta$ represent a better agreement with human evaluation, indicating a better evaluator. Abbreviations adopted: Cog. for CogVLM2-19B; Mini. for MiniCPM2.6; Qwen72B for QwenVL2-72B. \textcolor{red}{Pic.} represents adding image as input when evaluating.}
\label{evaluator}
\begin{center}
\renewcommand{\arraystretch}{1.5}
\setlength{\tabcolsep}{2.3pt}
\small
\begin{tabular}{llllllllll}
\hline
   Model & GPT-3.5Turbo & GPT-4Turbo & o1-preview & GPT-4o(\textcolor{red}{Pic.}) & Cog.(\textcolor{red}{Pic.})  & Mini.(\textcolor{red}{Pic.}) & Qwen72B(\textcolor{red}{Pic.})
\\ \hline 
Sp.~($\uparrow$) & 0.498 &0.711 &0.592 &0.720(\textcolor{red}{0.738})  &0.049(\textcolor{red}{0.027})   &0.048(\textcolor{red}{0.162})  &0.688(\textcolor{red}{0.678}) \\
~$\Delta$~~($\downarrow$) &1.39  &0.93 &1.14  &0.90(\textcolor{red}{0.84})  &2.18(\textcolor{red}{2.27})   &2.07(\textcolor{red}{1.88})  &1.02(\textcolor{red}{1.01})   \\
\hline
\end{tabular}
\end{center}
\end{table}

\begin{table}[]
\centering
\renewcommand{\arraystretch}{1.2}
\setlength{\tabcolsep}{10pt}
\caption{The Spearman's rank correlation coefficient~(Sp.) and average absolute differences~($\Delta$) between the evaluation scores of GPT-4o and human-annotated scores. Larger Sp. and smaller $\Delta$ represent a better agreement with human evaluation, indicating a better evaluator. Objective and Subjective denote objective and subjective questions. w and w/o represent evaluating with and without image.}
\label{sub_obj_evaluator}
\begin{tabular}{cccc|cccc}
\hline
 \multicolumn{4}{c|}{Objective}  & \multicolumn{4}{c}{Subjective}  \\                      \cline{1-4} \cline{5-8} 
          \multicolumn{2}{c}{w} & \multicolumn{2}{c|}{w/o} & \multicolumn{2}{c}{w} & \multicolumn{2}{c}{w/o} \\
          \redcline{1-2} \greencline{3-4} \bluecline{5-6} \greencline{7-8} 
                Sp.      &  $\Delta$            & Sp.          &$\Delta$           & Sp.         & $\Delta$           & Sp.      &$\Delta$    \\ \hline
                0.835 &0.57 &0.849  &0.60 &0.515 &1.23 &0.471 &1.3 \\
                \hline
\end{tabular}
\end{table}

\section{Conclusion}
In this paper, we have introduced MME-Finance, a pioneering effort to establish a bilingual multimodal benchmark tailored to evaluate the capabilities of MLLMs within the specialized financial domain. 
By encompassing a diverse range of financial open-ended questions, MME-Finance presents challenges that span from basic visual understanding to complex financial reasoning and expert-level decision-making.
Moreover, a novel evaluation strategy is proposed to ensure accurate evaluation of the MLLMs. Our detailed evaluation of 19 MLLMs shows that both open-source and proprietary models have significant limitations in handling complex financial questions. MME-Finance can serve as a critical tool to guide the development of MLLM capabilities in the financial domain. In future work, we plan to expand the data size of MME-Finance and integrate multi-turn dialogue scenarios for more comprehensive evaluation.

\bibliography{refs}
\bibliographystyle{bibstyle}

\clearpage
\appendix
\section{Appendix}
Statement: This paper is limited to academic research, and the OpenAI's products used do not violate the company's commercial regulations.

In this appendix, we provide further details regarding the proposed MME-Finance. ~\ref{subsec: chinese version} introduces the statistical overview and the experimental results of the Chinese version of MME-Finance. ~\ref{subsec: Other Results} presents experimental results with the prompt to allow ``Not Applicable'' response across all types of tasks. ~\ref{subsec: hard examples} provides some samples demonstrating the difficulty of recognizing mobile photos and the hallucination problems of MLLMs. ~\ref{subsec:infer_eval_prompt} includes the detailed inference and evaluation prompts. ~\ref{subsec: examples in tasks} exhibits example for each task.

\subsection{Chinese Version of MME-Finance.}\label{subsec: chinese version}

\begin{figure}[h]
    \centering
    \begin{subfigure}[b]{0.48\textwidth} 
        \centering
        \includegraphics[width=\textwidth]{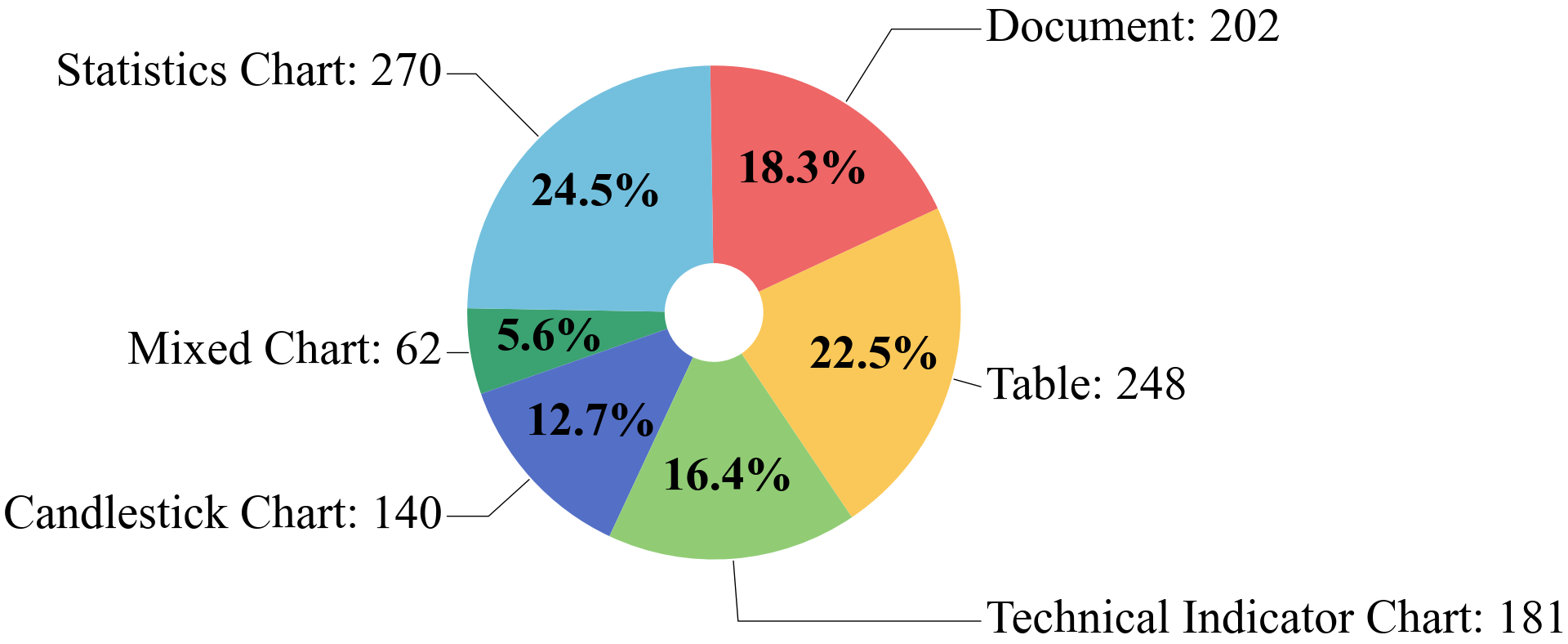} 
        \caption{}
        \label{fig:type_zh}
    \end{subfigure}
    \hspace{0.05\textwidth}
    \begin{subfigure}[b]{0.45\textwidth} 
        \centering
        \includegraphics[width=\textwidth]{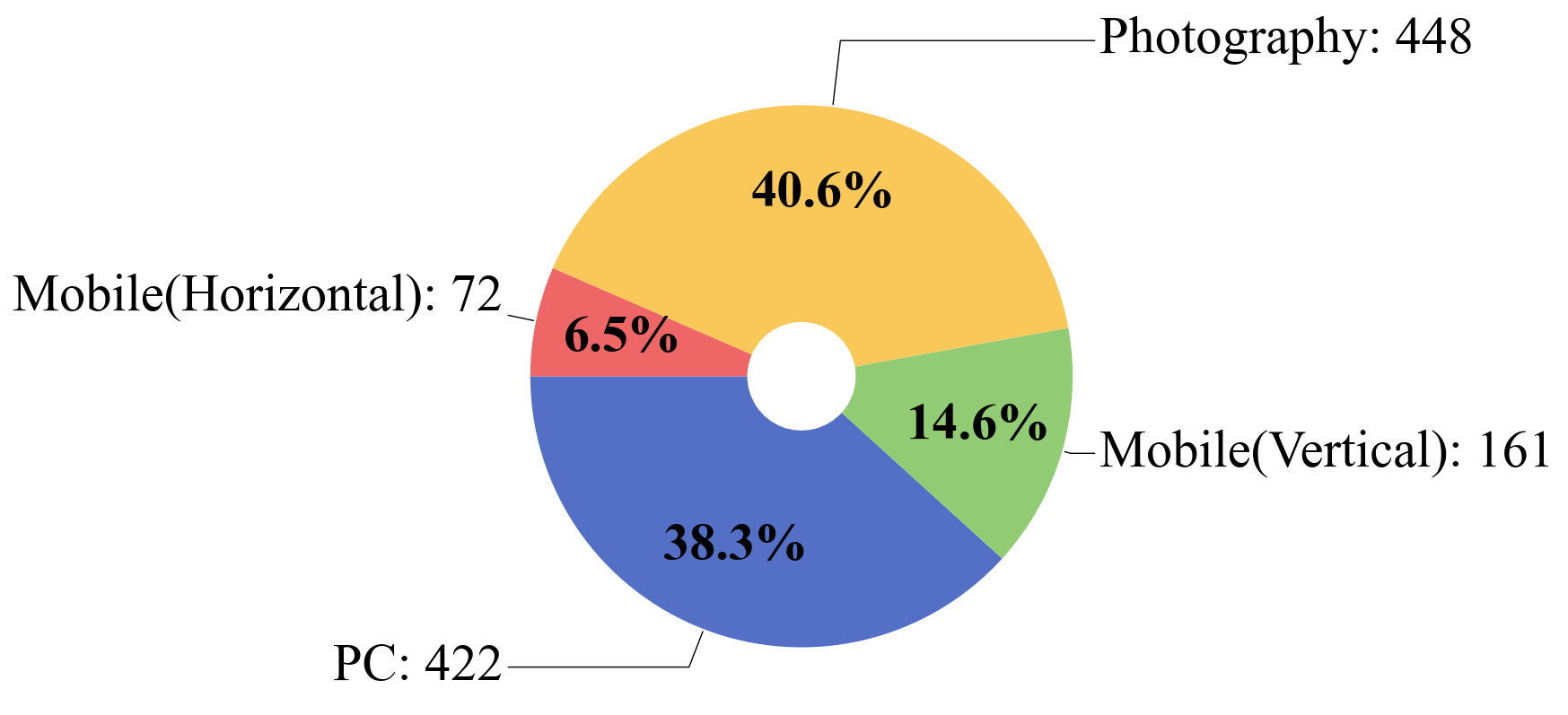} 
        \caption{}
        \label{fig:style_zh}
    \end{subfigure}
    \caption{Distribution of different (a) types and (b) styles of images of the Chinese version of MME-Finance.}
    \label{fig:mainfig_zh}
    \vspace{-3mm}
\end{figure}

\begin{wraptable}{r}{0.5\textwidth}
\vspace{-5mm}
\caption{Statistic of the number of samples in different capabilities and tasks of the Chinese version of MME-Finance.}
\label{tab:stat_zh}
\centering
\setlength{\tabcolsep}{2pt}
\begin{tabular}{lc}
\hline
 \textbf{Statistic} & \textbf{Number}
\\ \hline 
\textbf{Perception} &  640  \\
~- Image Caption &144  \\
~- OCR &182  \\
~- Entity Recognition &148 \\
~- Spatial Awareness  &166\\ \hline
\textbf{Reasoning} & 158 \\
~- Accurate Numerical Calculation & 126 \\
~- Estimated Numerical Calculation & 32 \\ \hline
\textbf{Cognition} & 285 \\ 
~- Risk Warning & 37 \\
~- Investment Advice & 91 \\
~- Reason Explanation & 13 \\
~- Financial Question Answer & 144 \\ \hline
\textbf{Hallucination } & 20 \\
~- Not Applicable &20 \\
\hline
\end{tabular}
\end{wraptable}

As shown in Table~\ref{tab:stat_zh}, Chinese MME-Finance contains 1,103 image-question-answer pairs, and has the same task categories as the English version.  
The number of samples per task varies from 13 to 182, with the ``OCR" task containing the most and ``Reason Explaination" the fewest. 
The distribution of the 6 image types and 4 image styles are shown in the Figure~\ref{fig:mainfig_zh}(\subref{fig:type_zh}) and Figure~\ref{fig:mainfig_zh}(\subref{fig:style_zh}), respectively. Statistics Charts have the largest proportion, while mixed charts are the least. As for the styles, photography has the largest number. 

Table~\ref{tasks_zh} shows the results of various open source MLLMs on Chinese MME-Finance in every task with Qwen2VL-72B evaluator. Among these models, Qwen2VL-72B achieves the best overall performance with 73.35\% accuracy, excelling in most tasks. The performance of Qwen2VL-7B is slightly lower, and the third is InternVL2-76B. The law of larger language model has a better performance is also applicable for the InternVL and the Qwen2VL series. The evaluation results with Qwen2VL-72B evaluator on Chinese MME-Finance for different types and styles of images are shown in Table~\ref{Types_styles_zh}. Qwen2VL-72B achieves the best results in all categories, followed by Qwen2VL-7B and InternVL2-76B. It is obviously that these models have lowest accuracy in the type of candlestick chart. As the styles of images, the mobile photograph is most difficult for most MLLMs. 

\begin{table}[h]
\caption{Evaluation results on the Chinese MME-Finance for all tasks. Abbreviations and color settings are the same as before. All numbers are denoted in \% with the max value of 100\%.}
\label{tasks_zh}
\begin{center}
\renewcommand{\arraystretch}{1.2}
\setlength{\tabcolsep}{2.6pt}
\footnotesize
\begin{tabular}{ccccccccccccc}
\hline
\multirow{2}{*}{\textbf{Model}} & \multirow{2}{*}{\textbf{Overall}} & \multicolumn{4}{c}{\textbf{Perception}}    & \multicolumn{2}{c}{\textbf{Reasoning}}  & \multicolumn{4}{c}{\textbf{Cognition}} & \multirow{2}{*}{\textbf{NA}}\\ 
\redcline{3-6} \greencline{7-8} \bluecline{9-12}
& &\textbf{IC} & \textbf{OCR} & \textbf{ER} & \textbf{SA}  & \textbf{ANC} & \textbf{ENC} & \textbf{RW} & \textbf{IA} & \textbf{RE} & \textbf{FQA}
\\ \hline 
\multicolumn{13}{c}{Open source MLLMs} \\
Yi-VL-34B &23.50  &43.89  &0.66  &9.86  &4.94    &23.97  &18.13  &20.00  &28.79 &60.00 &51.81 &\cellcolor{orange!40}100.0\\
CogVLM2-19B &35.32  &55.69  &41.10  &37.84  &16.02    &39.37  &29.38  &8.11  &31.43 &26.15 &28.47 &85.00\\
InternVL2-2B &39.76  &60.83  &54.51  &33.24  &16.02    &28.25  &19.38  &32.43  &44.40  &55.38 &44.86 &50.00 \\
InternVL2-4B &45.78  &67.22  &60.22  &43.51  &19.28  &51.43  &38.75  &31.89  &46.59  &63.08  &36.94  &47.00 \\
InternVL2-8B &58.44  &\cellcolor{green!40}73.47  &\cellcolor{green!40}76.92  &55.14  &25.18  &52.84  &42.50 &53.51  &\cellcolor{green!40}61.32  &\cellcolor{orange!40}76.92  &\cellcolor{cyan!40}67.78  &60.00 \\
InternVL2-76B &\cellcolor{green!40}62.63  &\cellcolor{green!40}73.47  &75.71 &\cellcolor{green!40}61.35  &\cellcolor{cyan!40}38.43  &\cellcolor{green!40}64.13  &\cellcolor{green!40}53.13 &\cellcolor{cyan!40}58.38  &\cellcolor{cyan!40}63.08  &\cellcolor{cyan!40}75.38  &\cellcolor{green!40}67.36  &45.00 \\
LLaVA-NEXT-7B &21.45  &50.69  &8.35  &12.16  &9.28  &16.03  &13.13 &12.43  &28.35  &46.15  &25.14  &\cellcolor{green!40}90.00 \\
LLaVA-NEXT-13B &19.87  &49.58  &8.68  &12.30  &13.01  &14.60  &9.38 &8.11  &24.84  &13.85  &17.64  &\cellcolor{green!40}90.00 \\
MiniCPM2.6 &38.60  &53.47  &64.29  &45.27  &23.98   &18.41  &27.50  &32.43  &36.70  &35.38 &27.92  &14.00 \\
Phi3-Vision &31.91  &57.92  &32.31  &40.68  &16.02   &29.05  &23.13  &22.70  &32.31 &43.08 &14.31 &75.00\\
Phi3.5-Vision &30.12  &55.97  &19.45  &20.27  &23.85   &20.48  &24.38  &28.65  &41.98 &41.54 &26.94 &\cellcolor{orange!40}100.0\\
Qwen2VL-2B &49.12  &65.97  &63.41  &48.38  &24.94   &39.05  &36.88  &36.76  &46.37  &56.92  &51.53 &\cellcolor{orange!40}100.0 \\
Qwen2VL-7B &\cellcolor{cyan!40}64.91  &\cellcolor{cyan!40}73.61  &\cellcolor{cyan!40}84.95  &\cellcolor{cyan!40}64.05  &\cellcolor{green!40}34.34   &\cellcolor{cyan!40}69.68  &\cellcolor{cyan!40}58.13  &\cellcolor{green!40}55.14  &59.34  &\cellcolor{green!40}67.69  &65.97 &\cellcolor{cyan!40}95.00 \\
Qwen2VL-72B &\cellcolor{orange!40}73.35  &\cellcolor{orange!40}79.58  &\cellcolor{orange!40}89.67  &\cellcolor{orange!40}73.24  &\cellcolor{orange!40}55.90   &\cellcolor{orange!40}73.81  &\cellcolor{orange!40}73.13  &\cellcolor{orange!40}69.19  &\cellcolor{orange!40}65.05  &\cellcolor{orange!40}76.92  &\cellcolor{orange!40}74.17 &60.00 \\ \hline
\end{tabular}
\end{center}
\end{table}

\begin{table}[t]
\caption{Evaluation results on Chinese MME-Finance for different types and styles of images. Abbreviations and color settings are the same as before. All numbers are denoted in \% with the max value of 100\%.}
\label{Types_styles_zh}
\begin{center}
\renewcommand{\arraystretch}{1.2}
\setlength{\tabcolsep}{5pt}
\small
\begin{tabular}{lllllll|llll}
\hline
\textbf{Model} &\textbf{Candle.} & \textbf{Tech.} & \textbf{Stat.} & \textbf{Tab.} & \textbf{Doc.} & \textbf{Mixed} & \textbf{CS} & \textbf{MP} & \textbf{VS} & \textbf{HS} 
\\ \hline 
\multicolumn{11}{c}{Open source MLLMs} \\
Yi-VL-34B &26.00  &21.66  &22.87  &23.55  &23.63  &24.52  &26.30  &20.49  &25.09  &22.22\\
CogVLM2-19B &38.86  &33.59  &31.19  &37.74  &37.48  &37.48  &37.11  &32.50  &36.77  &39.17\\
InternVL2-2B &36.29  &35.80  &39.01  &37.90  &48.22  &32.26  &43.70  &36.92  &39.63  &34.72  \\
InternVL2-4B &34.43  &44.20  &46.93  &48.31  &51.56  &37.10  &51.90  &40.80  &43.72  &45.56  \\
InternVL2-8B &49.71  &55.69  &56.44  &59.60  &66.44  &53.23  &61.80  &55.31 &62.11  &50.00  \\
InternVL2-76B &\cellcolor{cyan!40}55.86  &\cellcolor{cyan!40}64.75  &\cellcolor{green!40}61.39  &\cellcolor{green!40}62.74  &\cellcolor{green!40}67.56  &\cellcolor{green!40}53.87  &\cellcolor{green!40}65.55  &\cellcolor{green!40}57.28 &\cellcolor{cyan!40}69.44  &\cellcolor{cyan!40}63.61  \\
LLaVA-NEXT-7B &29.57  &20.88  &20.00  &16.21  &25.33  &13.55  &22.89 &19.42  &23.23  &21.67 \\
LLaVA-NEXT-13B &24.14  &21.66  &21.39  &16.37  &19.56  &15.48  &20.33 &19.29  &19.88  &20.83 \\
MiniCPM2.6 &36.57  &36.69  &37.92  &40.08  &44.81  &18.06  &38.58  &35.80 &47.33 & 36.67   \\
Phi3-Vision &31.71  &35.47  &29.21  &31.45   &34.30  &22.26   &36.45  &27.32 &31.55 &34.72\\
Phi3.5-Vision &30.29  &35.03  &27.92  &25.48   &33.33  &27.10   &32.09  &27.54 &29.81 &35.28\\
Qwen2VL-2B &40.71  &42.10  &49.11  &49.68  &59.41  &41.61   &49.95  &48.88  &51.80  &39.72   \\
Qwen2VL-7B &\cellcolor{green!40}55.71  &\cellcolor{green!40}60.55  &\cellcolor{cyan!40}69.21  &\cellcolor{cyan!40}66.13  &\cellcolor{cyan!40}68.37  &\cellcolor{cyan!40}64.52  &\cellcolor{cyan!40}69.29  &\cellcolor{cyan!40}62.41  &\cellcolor{green!40}62.73  &\cellcolor{green!40}59.72  \\
Qwen2VL-72B &\cellcolor{orange!40}64.14  &\cellcolor{orange!40}71.71  &\cellcolor{orange!40}77.52  &\cellcolor{orange!40}75.65  &\cellcolor{orange!40}75.26  &\cellcolor{orange!40}67.74  &\cellcolor{orange!40}76.35  &\cellcolor{orange!40}69.96  &\cellcolor{orange!40}74.53  &\cellcolor{orange!40}74.17 \\ \hline
\end{tabular}
\end{center}
\end{table}

\subsection{Results of NA Prompt For ALL Tasks.}\label{subsec: Other Results}
Table~\ref{na_tasks} and Table~\ref{NA_Types_formats} shows the performance of MLLMs with the prompt to allow ``Not Applicable'' response across all types of tasks. It is clear that most models have a lower performance in the setting, which means the hallucination problem is quite common in MLLMs. Although some models have a high recall of the ``Not Applicable'' question, their overall accuracy is low. It shows that these models tend to answer ``Not Applicable'' for some unsure questions.

\begin{table}[h]
\caption{Evaluation results on MME-Finance with the prompt to allow “Not Applicable” response across all types of tasks. Abbreviations and color settings are the same as before. All numbers are denoted in \% with the max value of 100\%.}
\label{na_tasks}
\begin{center}
\renewcommand{\arraystretch}{1.2}
\setlength{\tabcolsep}{2.6pt}
\footnotesize
\begin{tabular}{ccccccccccccc}
\hline
\multirow{2}{*}{\textbf{Model}} & \multirow{2}{*}{\textbf{Overall}} & \multicolumn{4}{c}{\textbf{Perception}}    & \multicolumn{2}{c}{\textbf{Reasoning}}  & \multicolumn{4}{c}{\textbf{Cognition}} & \multirow{2}{*}{\textbf{NA}}\\ 
\redcline{3-6} \greencline{7-8} \bluecline{9-12}
& &\textbf{IC} & \textbf{OCR} & \textbf{ER} & \textbf{SA}  & \textbf{ANC} & \textbf{ENC} & \textbf{RW} & \textbf{IA} & \textbf{RE} & \textbf{FQA}
\\ \hline 
\multicolumn{13}{c}{Open source MLLMs} \\
CogVLM2-19B &31.24  &36.22  &42.02  &26.99  &7.95    &34.14  &19.52  &13.64  &38.11 &32.22 &48.44 &70.91\\
InternVL2-2B &32.16  &61.22  &32.81  &12.76  &4.72    &32.78  &21.43  &54.55  &47.92  &63.33 &50.20 &50.00 \\
InternVL2-4B &45.93  &68.05  &54.27  &28.22  &18.69  &54.74  &32.38  &68.18  &50.19  &68.89  &59.46  &59.09 \\
InternVL2-8B &50.59  &70.00  &60.11  &33.99  &\cellcolor{green!40}23.84  &\cellcolor{green!40}62.56  &\cellcolor{cyan!40}40.00 &\cellcolor{cyan!40}76.36  &60.75  &\cellcolor{green!40}75.56  &58.37  &55.45 \\
LLaVA-NEXT-7B &20.10  &58.05  &6.18  &2.70  &1.31  &6.62  &3.33 &43.64  &37.74  &16.67  &43.27  &70.00 \\
MiniCPM2.6 &\cellcolor{green!40}48.37  &69.63  &\cellcolor{green!40}62.81  &\cellcolor{green!40}34.85  &21.31   &54.89  &28.57  &50.00  &40.38  &45.56 &62.31  &80.00 \\
Phi3-Vision &37.06  &69.51  &58.43  &29.57  &11.88   &22.11  &3.81  &7.27  &16.60 &41.11 &47.76 &\cellcolor{cyan!40}98.18\\
Phi3.5-Vision &28.69  &66.83  &33.03  &13.87  &8.12   &17.14  &6.19  &2.73  &15.47 &22.22 &45.58 &\cellcolor{green!40}96.36\\
Qwen2VL-7B &32.40  &62.32  &40.00  &8.10  &7.07   &41.80  &22.86  &33.64  &40.00  &10.00  &40.14 &\cellcolor{orange!40}100.0 \\
Qwen2VL-2B &32.47  &62.32  &40.45  &9.08  &6.72   &41.80  &25.24  &31.82  &36.23  &13.33  &40.14 &\cellcolor{orange!40}100.0 \\
Qwen2VL-72B &\cellcolor{orange!40}62.97  &\cellcolor{cyan!40}80.49  &\cellcolor{orange!40}83.26  &\cellcolor{orange!40}50.43  &\cellcolor{cyan!40}25.50   &\cellcolor{orange!40}78.95  &\cellcolor{orange!40}46.67  &\cellcolor{green!40}73.64  &\cellcolor{orange!40}68.30  &73.33  &\cellcolor{orange!40}73.06 &86.36 \\ \hline
\multicolumn{13}{c}{Proprietary MLLMs} \\
GPT-4o-5-13 &42.12  &\cellcolor{green!40}72.07  &26.74  &26.99  &19.56  &29.17  
&\cellcolor{green!40}38.57 &70.90  &\cellcolor{green!40}63.77  &\cellcolor{cyan!40}76.67  &\cellcolor{cyan!40}71.02  &72.72 \\
GPT-4o-mini &41.32 &63.54  &58.54  &28.34  &14.32   &24.06  &6.19  &52.73  &30.19  &63.33  &68.57 &\cellcolor{orange!40}100.0 \\
GPT-4o &\cellcolor{cyan!40}61.35  &\cellcolor{orange!40}83.66  &\cellcolor{cyan!40}78.54  &\cellcolor{cyan!40}46.38  &\cellcolor{orange!40}28.73  &\cellcolor{cyan!40}71.43  &\cellcolor{cyan!40}40.00 &\cellcolor{orange!40}80.00  &\cellcolor{cyan!40}65.66  &\cellcolor{orange!40}77.78  &\cellcolor{green!40}70.88  &80.00 \\ \hline
\end{tabular}
\end{center}
\end{table}

\begin{table}[t]
\caption{Evaluation results on MME-Finance for different types and formats of images with the prompt to allow “Not Applicable” response across all types of tasks. Abbreviations and color settings are the same as before. All numbers are denoted in \% with the max value of 100\%.}
\label{NA_Types_formats}
\begin{center}
\renewcommand{\arraystretch}{1.2}
\setlength{\tabcolsep}{5pt}
\small
\begin{tabular}{lllllll|llll}
\hline
\textbf{Model} &\textbf{Candle.} & \textbf{Tech.} & \textbf{Stat.} & \textbf{Tab.} & \textbf{Doc.} & \textbf{Mixed} & \textbf{CS} & \textbf{MP} & \textbf{VS} & \textbf{HS} 
\\ \hline 
\multicolumn{11}{c}{Open source MLLMs} \\
CogVLM2-19B &25.31  &19.89  &33.13  &32.52  &37.53  &39.05  &31.70  &29.66  &33.33  &35.84\\
InternVL2-2B &25.03  &25.57  &36.64  &32.37  &33.88  &53.33  &36.21  &28.82  &25.76  &29.09  \\
InternVL2-4B &30.91  &34.43  &54.35  &45.47  &52.37  &56.19  &49.68  &43.63  &39.39  &38.70  \\
InternVL2-8B &39.02  &38.18  &\cellcolor{green!40}55.27  &\cellcolor{green!40}50.72  &\cellcolor{cyan!40}58.08  &\cellcolor{green!40}69.52  &\cellcolor{green!40}54.98  &47.13 &\cellcolor{green!40}49.70  &41.04  \\
LLaVA-NEXT-7B &16.08  &18.75  &22.82  &19.31  &19.93  &33.33  &22.42 &16.71  &22.12  &22.60 \\
MiniCPM2.6 &\cellcolor{green!40}39.86  &\cellcolor{green!40}44.66  &51.53  &48.85  &52.23  &38.10  &47.26  &\cellcolor{green!40}48.48 &48.48 & \cellcolor{green!40}55.58   \\
Phi3-Vision &24.62  &28.18  &42.82  &40.00   &42.16  &12.38   &38.66  &35.86 &29.70 &39.22\\
Phi3.5-Vision &20.98  &22.95  &30.23  &30.29   &34.36  &10.48   &32.92  &22.95 &31.82 &30.91\\
Qwen2VL-2B &22.80  &24.32  &35.34  &37.63  &33.20  &48.57   &33.86  &30.42  &29.39  &36.62   \\
Qwen2VL-7B &23.92  &23.86  &35.57  &37.91  &33.06  &43.81  &33.54  &31.18  &29.39  &35.32  \\
Qwen2VL-72B &\cellcolor{cyan!40}52.31  &\cellcolor{orange!40}56.14  &\cellcolor{orange!40}69.16  &\cellcolor{cyan!40}64.32  &\cellcolor{orange!40}64.67  &\cellcolor{cyan!40}74.29  &\cellcolor{orange!40}65.78  &\cellcolor{orange!40}59.41  &\cellcolor{cyan!40}63.94  &\cellcolor{cyan!40}63.90 \\ \hline
\multicolumn{11}{c}{Proprietary MLLMs} \\
GPT-4o-5-13 &\cellcolor{green!40}39.86  &33.86  &53.13  &36.72  &40.43  &63.81  &45.67  &36.58  &49.39  &44.41  \\
GPT-4o-mini &26.15  &26.02  &49.90  &46.87  &43.88  &50.48  &45.99  &36.50  &35.76  &42.08  \\
GPT-4o &\cellcolor{orange!40}52.45  &\cellcolor{cyan!40}51.93  &\cellcolor{cyan!40}67.70  &\cellcolor{orange!40}68.79  &\cellcolor{green!40}57.12  &\cellcolor{orange!40}76.19  &\cellcolor{cyan!40}64.77  &\cellcolor{cyan!40}55.82  &\cellcolor{orange!40}66.67  &\cellcolor{orange!40}66.23  \\ \hline
\end{tabular}
\end{center}
\end{table}

\subsection{Hard Examples}\label{subsec: hard examples}
\begin{figure}[h]
\begin{center}
\includegraphics[width=0.9\columnwidth]{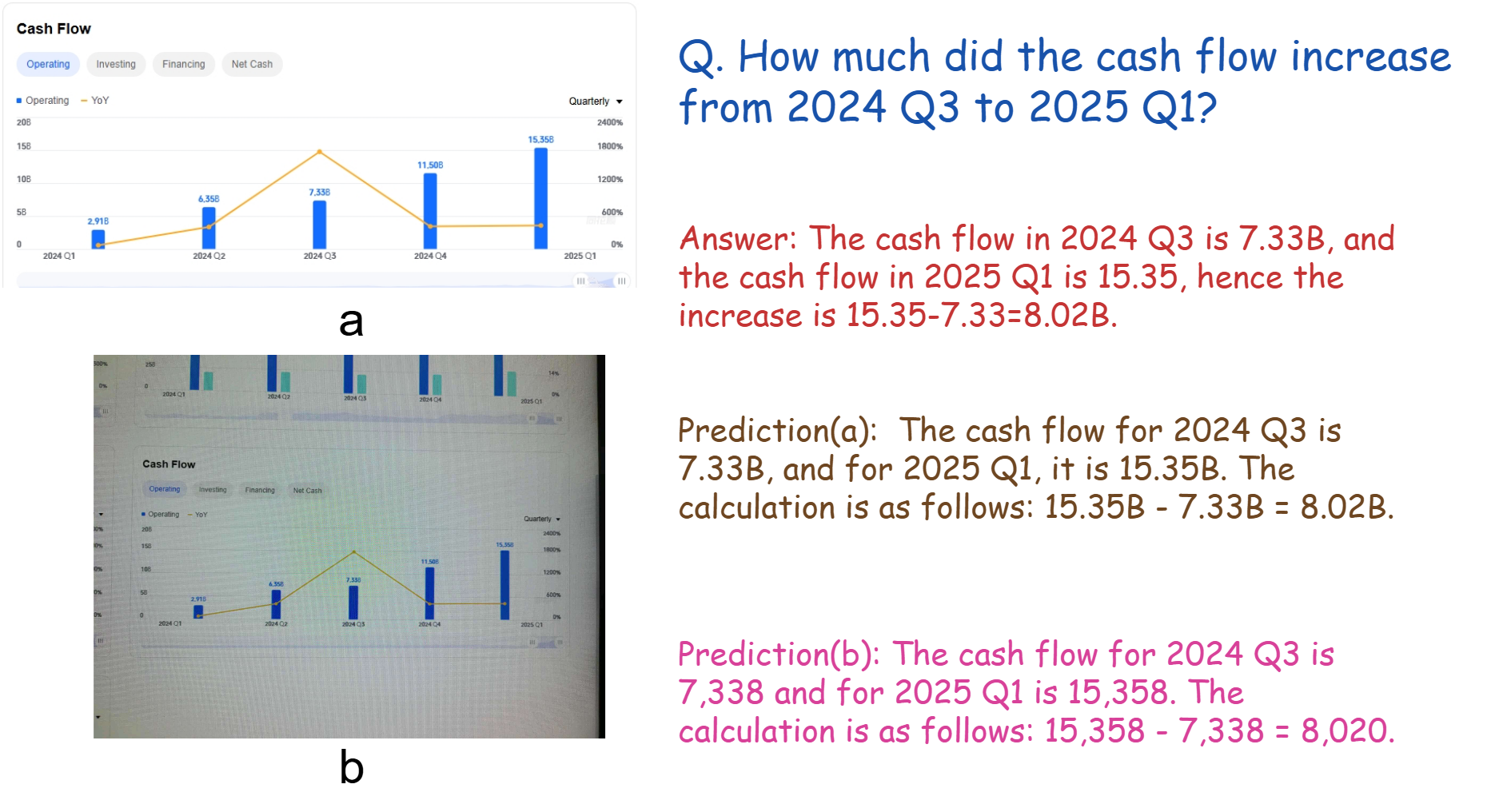}
\end{center}
\caption{Comparison of the difficulty of recognizing computer screenshot versus photos taken with a mobile phone.}
\label{fig: hard_computer_phone}
\end{figure}

\begin{figure}[h]
\begin{center}
\includegraphics[width=.8\columnwidth]{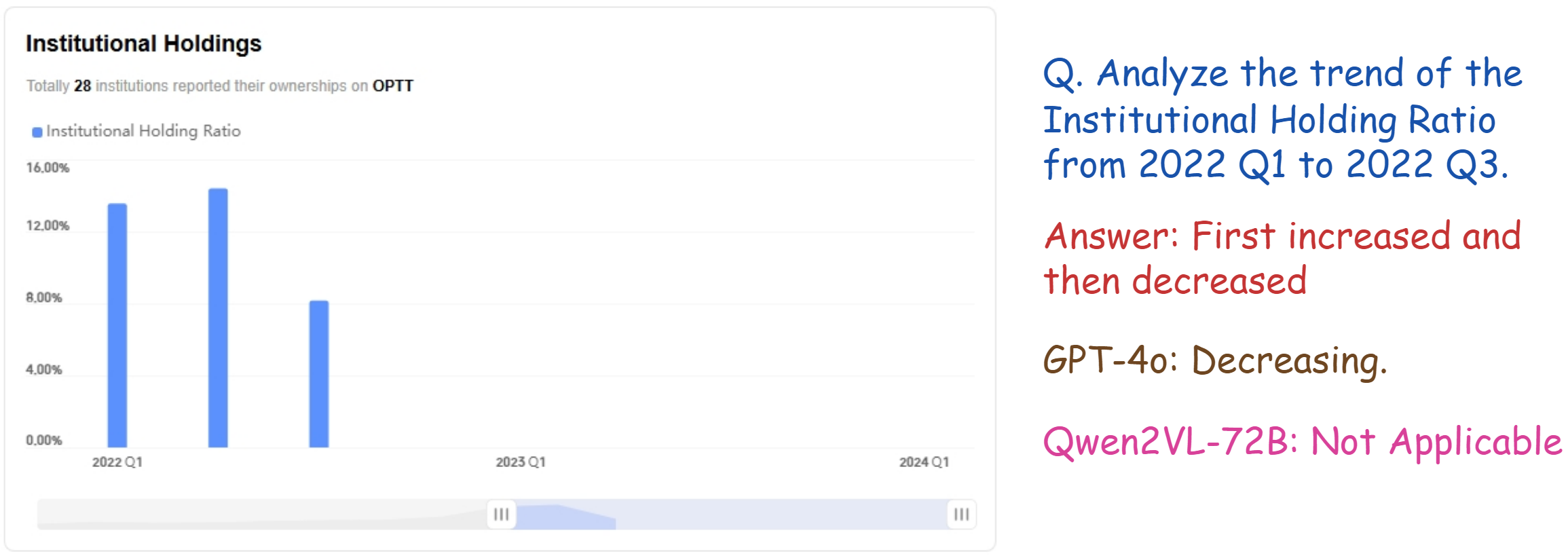}
\end{center}
\caption{The display of hallucination problems of MLLMs.}
\label{fig: hallu}
\end{figure}

In this section, we present some hard examples about the difficulty of mobile photos and hallucination problems of MLLMs. As shown in Figure~\ref{fig: hard_computer_phone}, the two questions have similar content. When feeding the two images into the same model, the responses are different. For the picture taken with a mobile phone, the model mistakenly identifies decimal points as commas and the letter B as the number 8. And the model accurately identifies corresponding elements in the computer screenshot. This indicates that the perception of mobile phone photos is a challenge for some MLLMs. Figure~\ref{fig: hallu} illustrates a example of the hallucination problem. GPT-4o cannot recognize the initial increase trend, while Qwen2VL-72B totally unable to perceive trends. 

\subsection{Inference and Evaluation Prompt.}\label{subsec:infer_eval_prompt}
\begin{figure}[p]
\begin{center}
\includegraphics[width=.95\columnwidth]{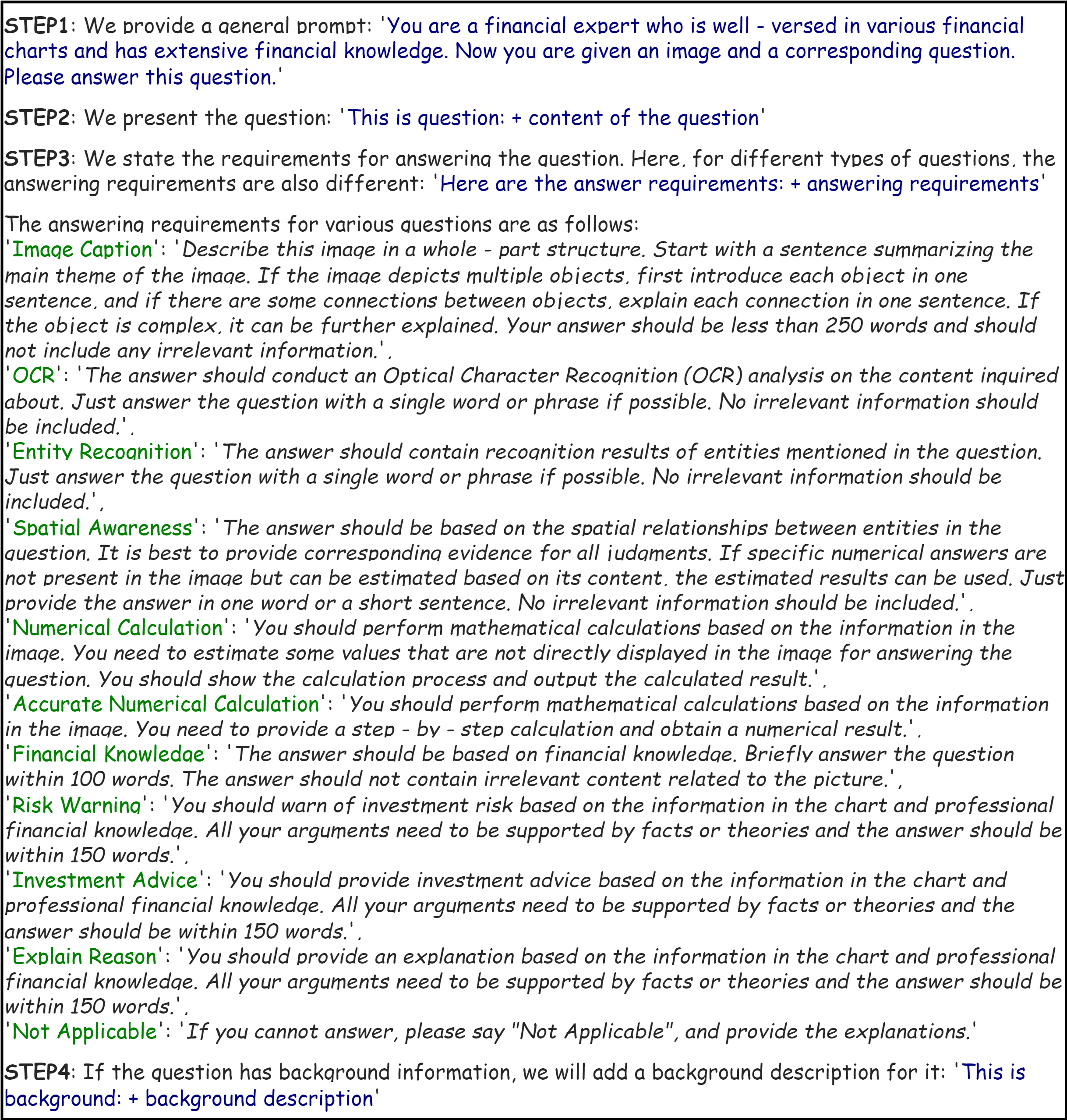}
\end{center}
\caption{Inference prompt.}
\label{fig: inference prompt}
\end{figure}

\begin{figure}[p]
\begin{center}
\includegraphics[width=.95\columnwidth]{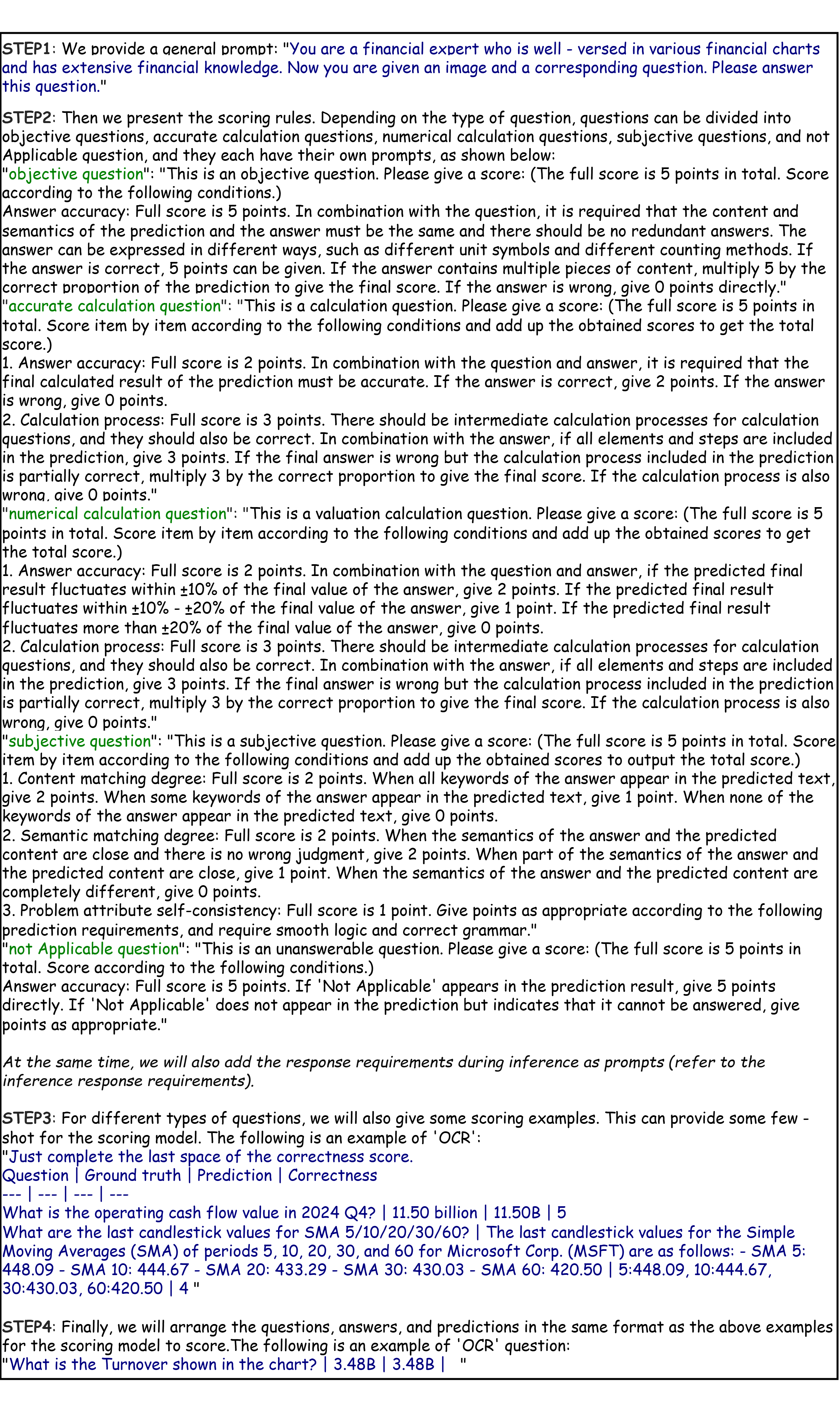}
\end{center}
\caption{Evaluation prompt.}
\label{fig: evaluation prompt.}
\end{figure}

Figure~\ref{fig: inference prompt} and Figure~\ref{fig: evaluation prompt.} shows the detailed inference prompt and evaluation prompt. 

\subsection{Definition and Example For Each Task.}\label{subsec: examples in tasks}
In this section, we provide a detailed definition of each task and present corresponding examples to help reader learn about these tasks. 

\textbf{Perception}
\begin{enumerate}
    \item \textbf{Image Caption:} Generate a textual description that accurately represents the content, context, and significant elements of an image.
    \item \textbf{OCR:} Recognition of text, number in the image.
    \item \textbf{Entity Recognition:} Recognition and understanding of visual elements(such as color, shape) or named entity in the image.
    \item \textbf{Spatial Awareness:} Understand the position and spatial relationship of the elements in the image.
\end{enumerate}

\textbf{Reasoning}
\begin{enumerate}
    \item \textbf{Accurate Numerical Calculation:} Perform accurate numerical calculation or numerical comparison based on the number in the image.
    \item \textbf{Estimated Numerical Calculation:} Obtain approximate values based on relevant clues(such as spatial location) and perform numerical calculation.
\end{enumerate}

\textbf{Cognition}
\begin{enumerate}
    \item \textbf{Risk Warning:} Give an investment risk description based on the information in the image (and background information).
    \item \textbf{Investment Advice:} Give an investment advice based on the information in the image (and background information).
    \item \textbf{Explain Reason:} Give an reason for the phenomenon indicated in the question based on the information in the image (and background information).
    \item \textbf{Financial Question Answer:} Answer objective financial questions based on the scored general financial knowledge.
\end{enumerate}

\begin{figure}[p]
\begin{center}
\includegraphics[width=0.9\columnwidth]{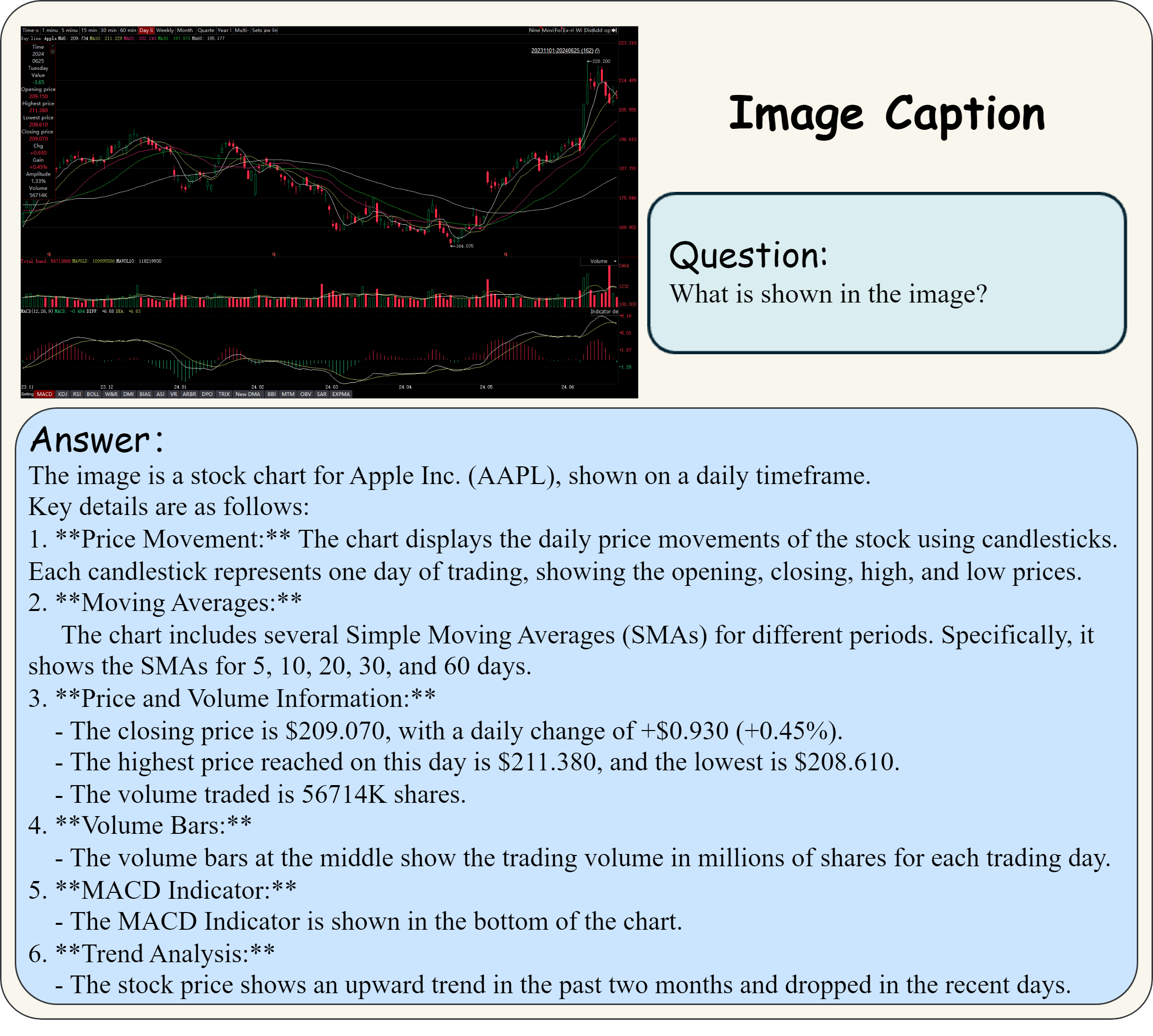}
\end{center}
\caption{Image Caption.}
\label{fig: Image_caption}
\end{figure}

\begin{figure}[p]
\begin{center}
\includegraphics[width=.8\columnwidth]{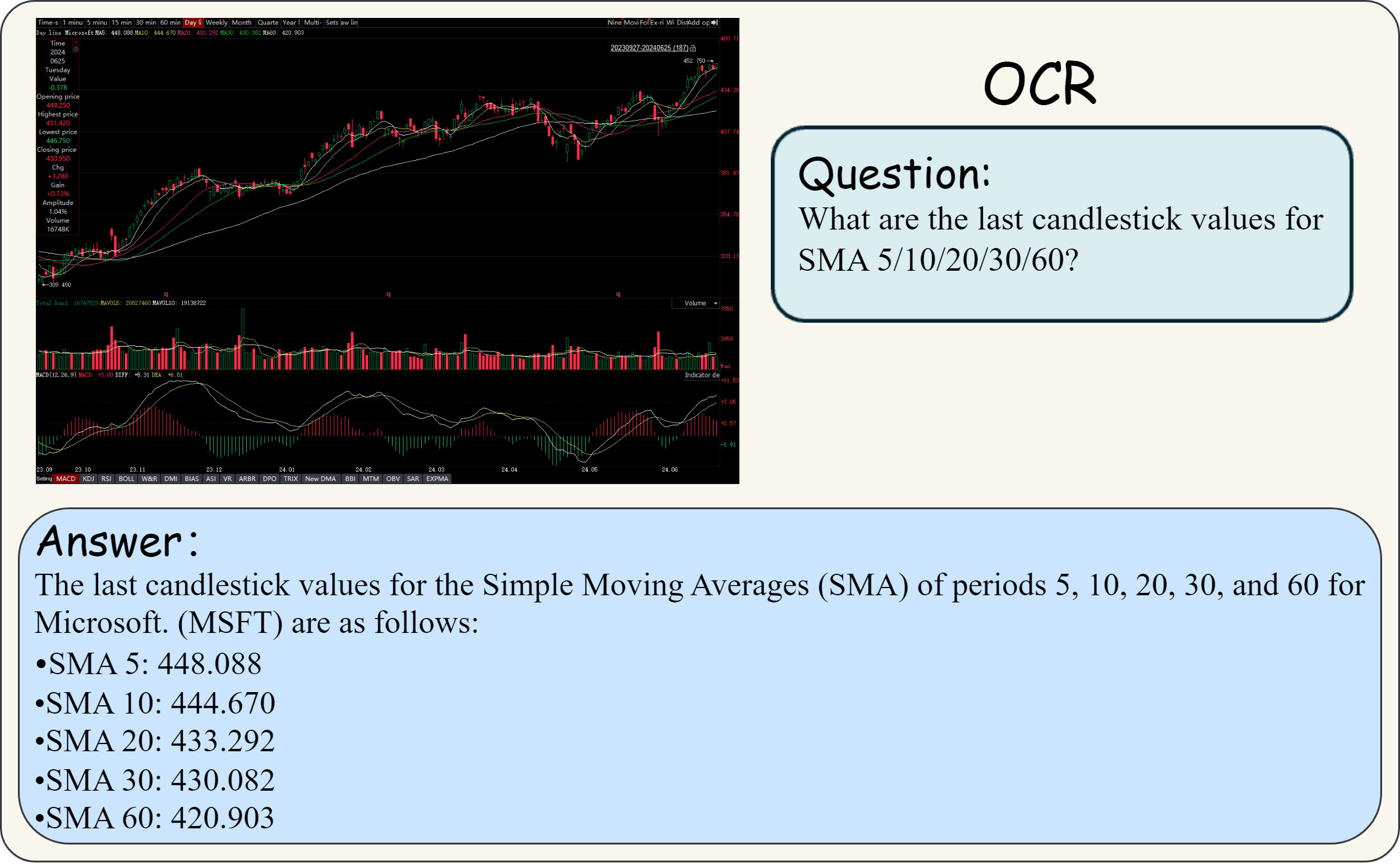}
\end{center}
\caption{OCR.}
\label{fig: OCR}
\end{figure}

\begin{figure}[h]
\begin{center}
\includegraphics[width=.8\columnwidth]{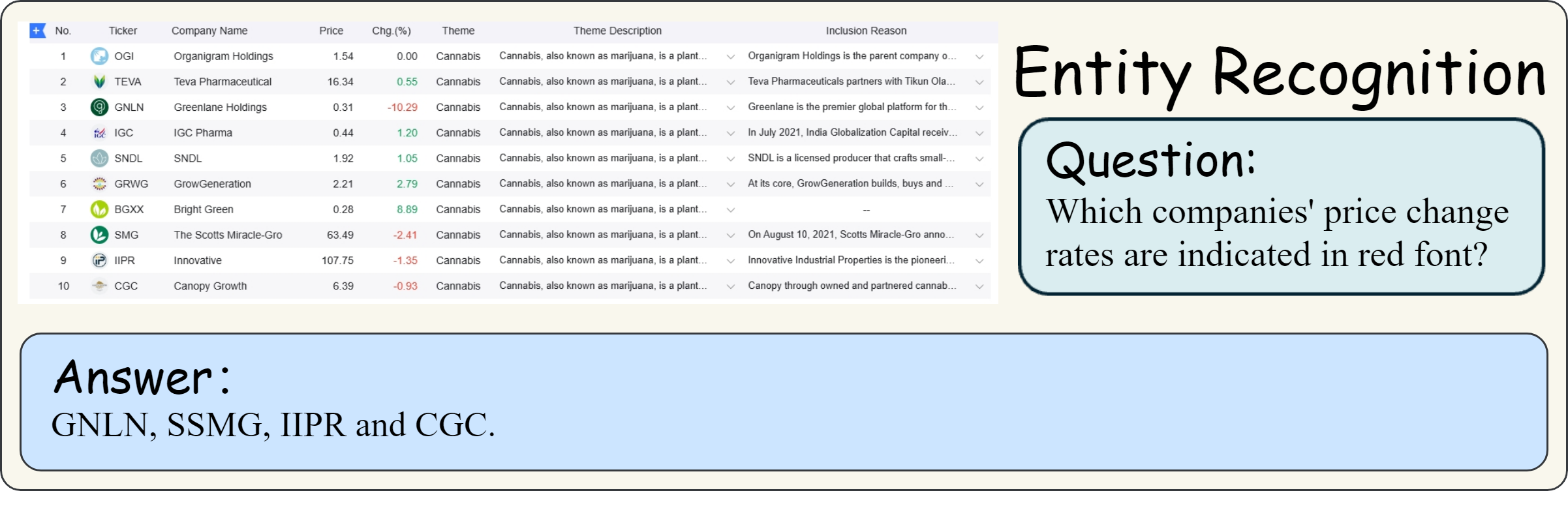}
\end{center}
\caption{Entity Recognition.}
\label{fig: entity_recognition}
\end{figure}

\begin{figure}[h]
\begin{center}
\includegraphics[width=.8\columnwidth]{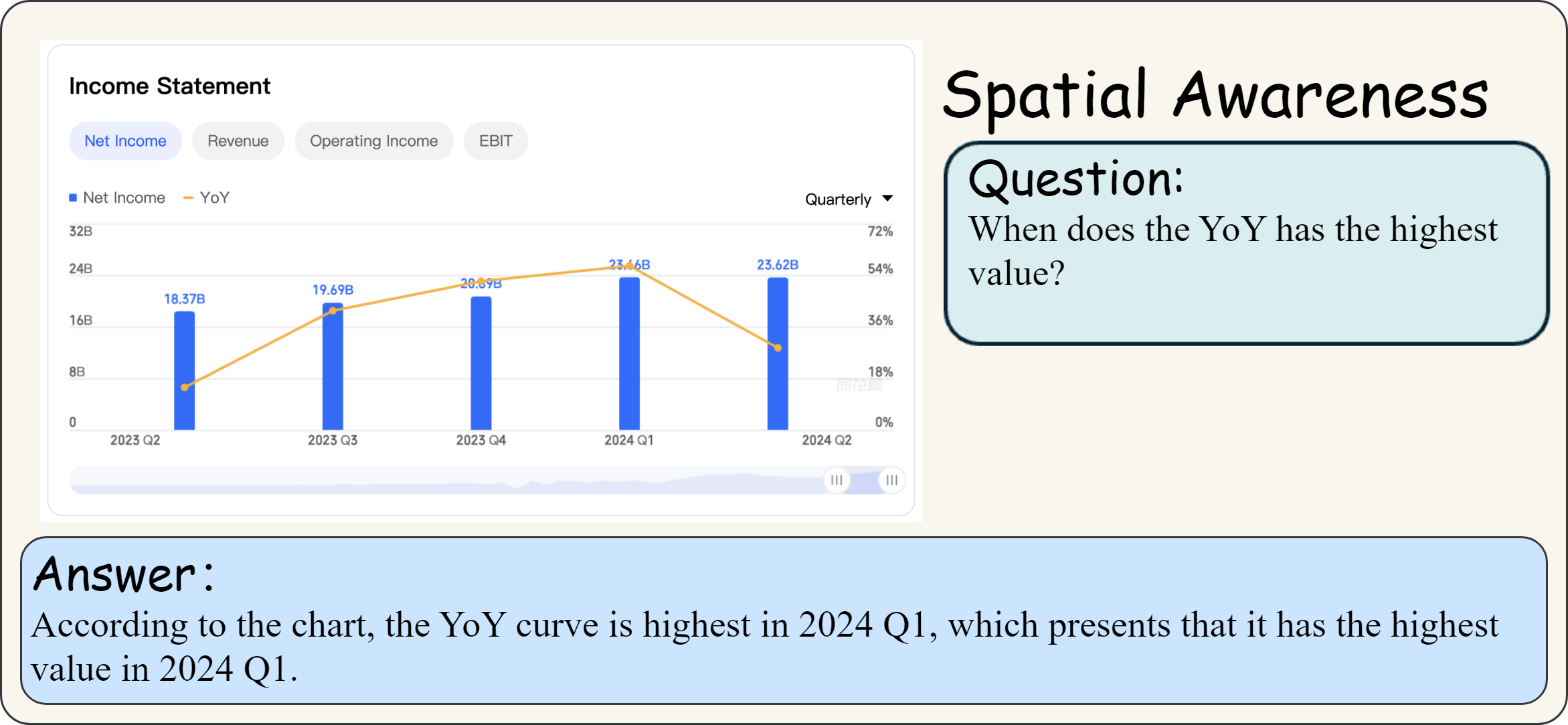}
\end{center}
\caption{Spatial Awareness.}
\label{fig: ENC}
\end{figure}

\begin{figure}[h]
\begin{center}
\includegraphics[width=.8\columnwidth]{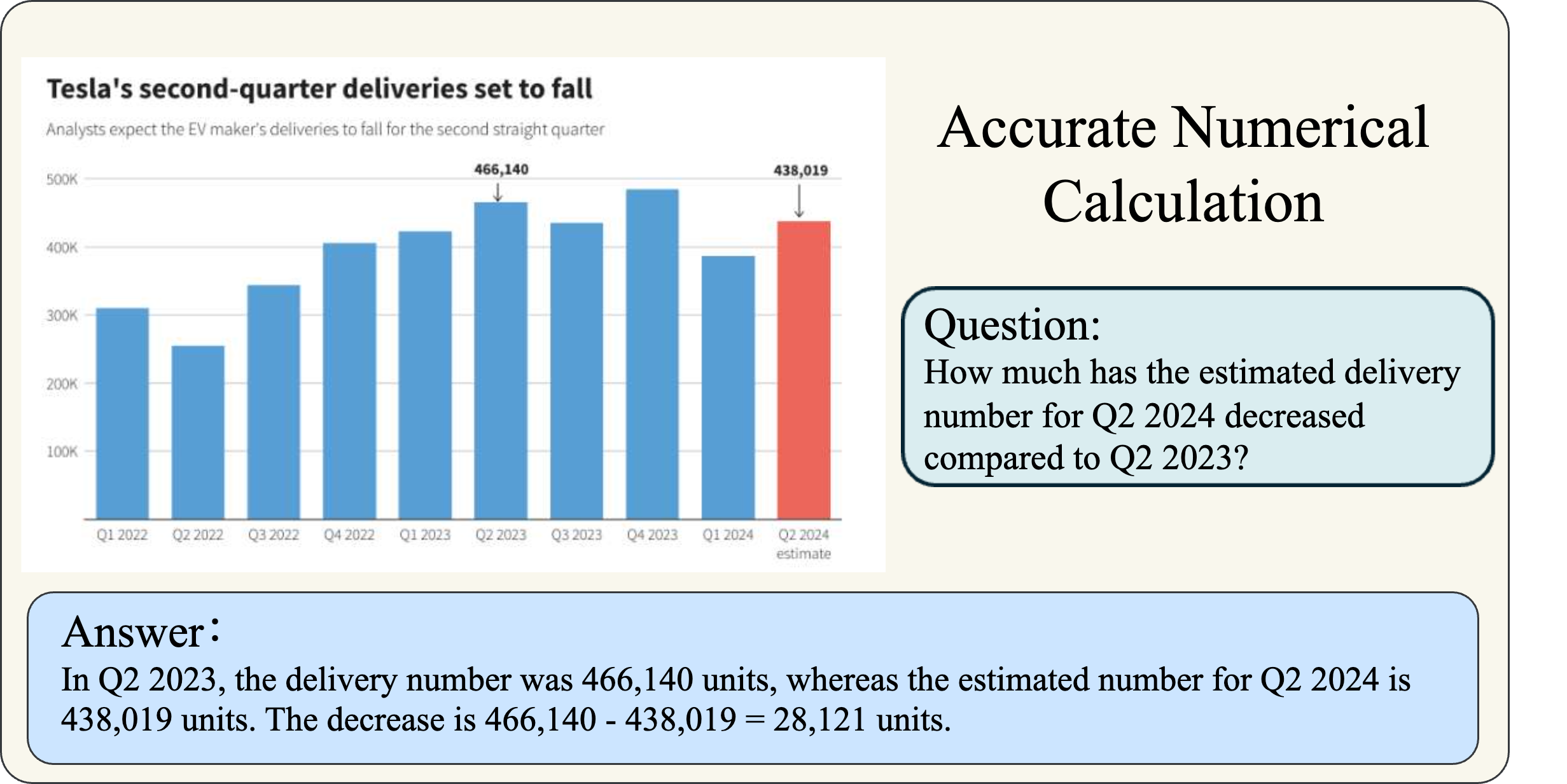}
\end{center}
\caption{Accurate Numerical Calculation.}
\label{fig: ENC}
\end{figure}

\begin{figure}[h]
\begin{center}
\includegraphics[width=.8\columnwidth]{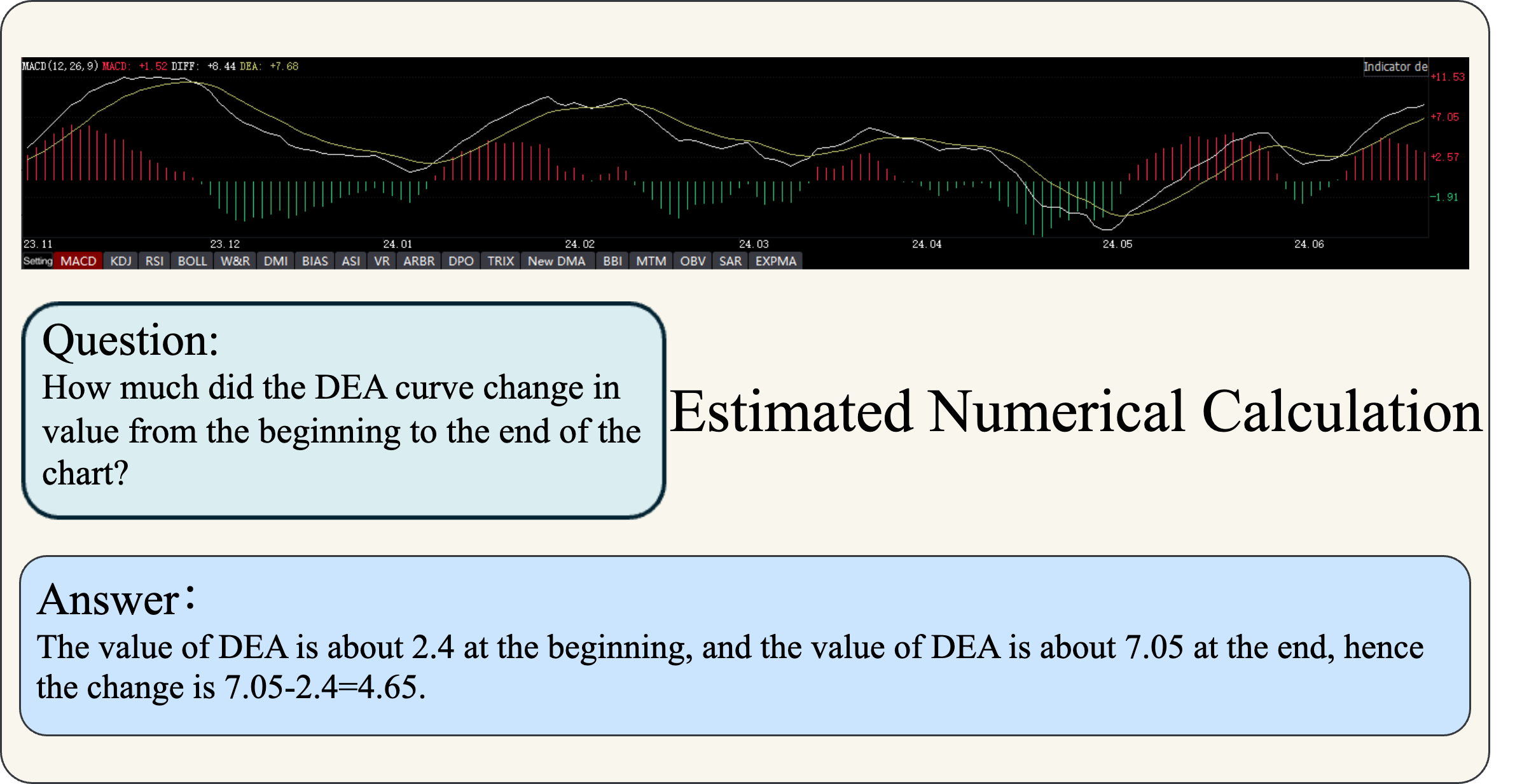}
\end{center}
\caption{Estimated Numerical Calculation.}
\label{fig: ENC}
\end{figure}

\begin{figure}[h]
\begin{center}
\includegraphics[width=.8\columnwidth]{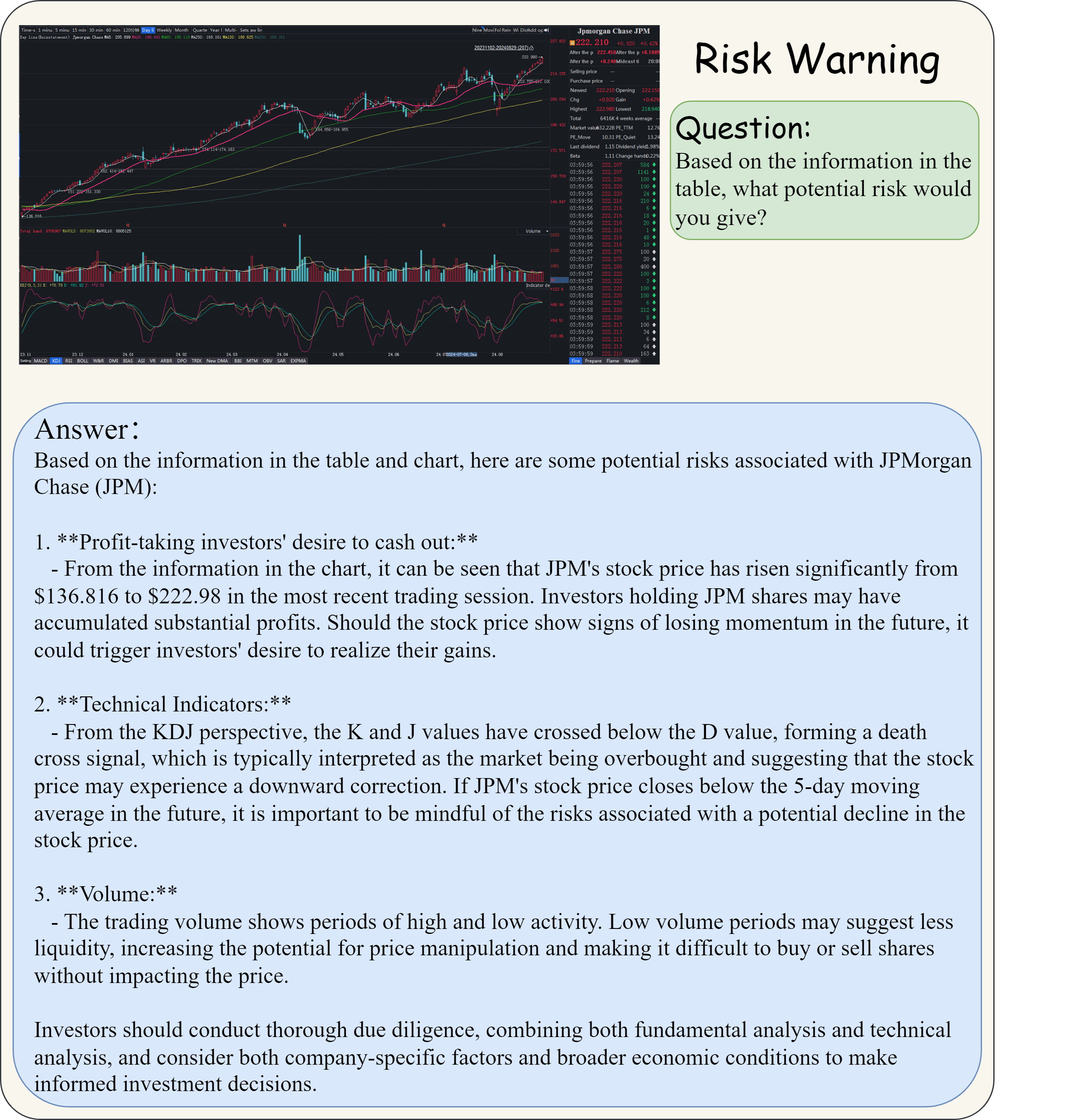}
\end{center}
\caption{Risk Warning.}
\label{fig: ENC}
\end{figure}

\begin{figure}[h]
\begin{center}
\includegraphics[width=.8\columnwidth]{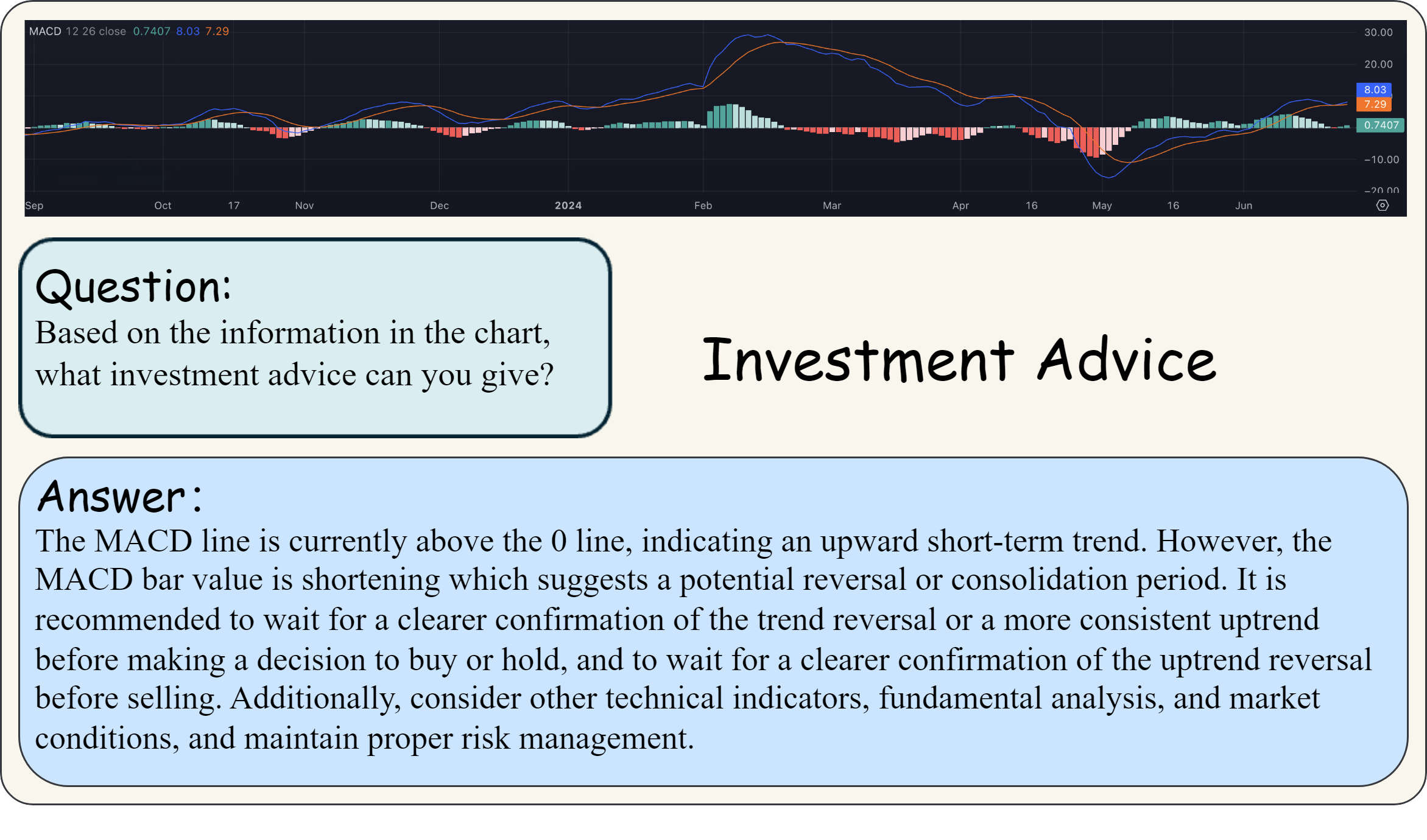}
\end{center}
\caption{Investment Advice.}
\label{fig: ENC}
\end{figure}

\begin{figure}[h]
\begin{center}
\includegraphics[width=.8\columnwidth]{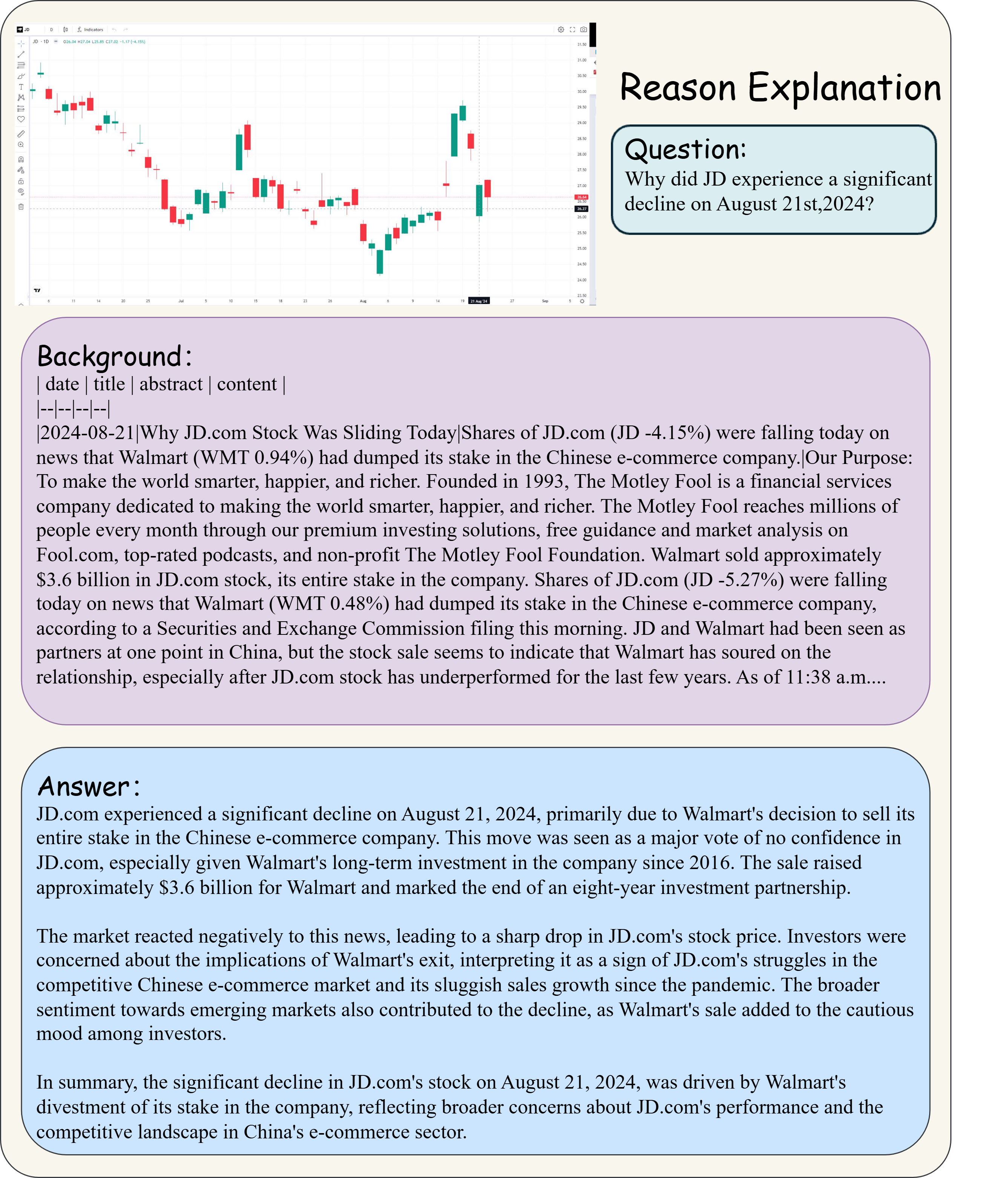}
\end{center}
\caption{Reason Explanation.}
\label{fig: ENC}
\end{figure}

\begin{figure}[h]
\begin{center}
\includegraphics[width=.8\columnwidth]{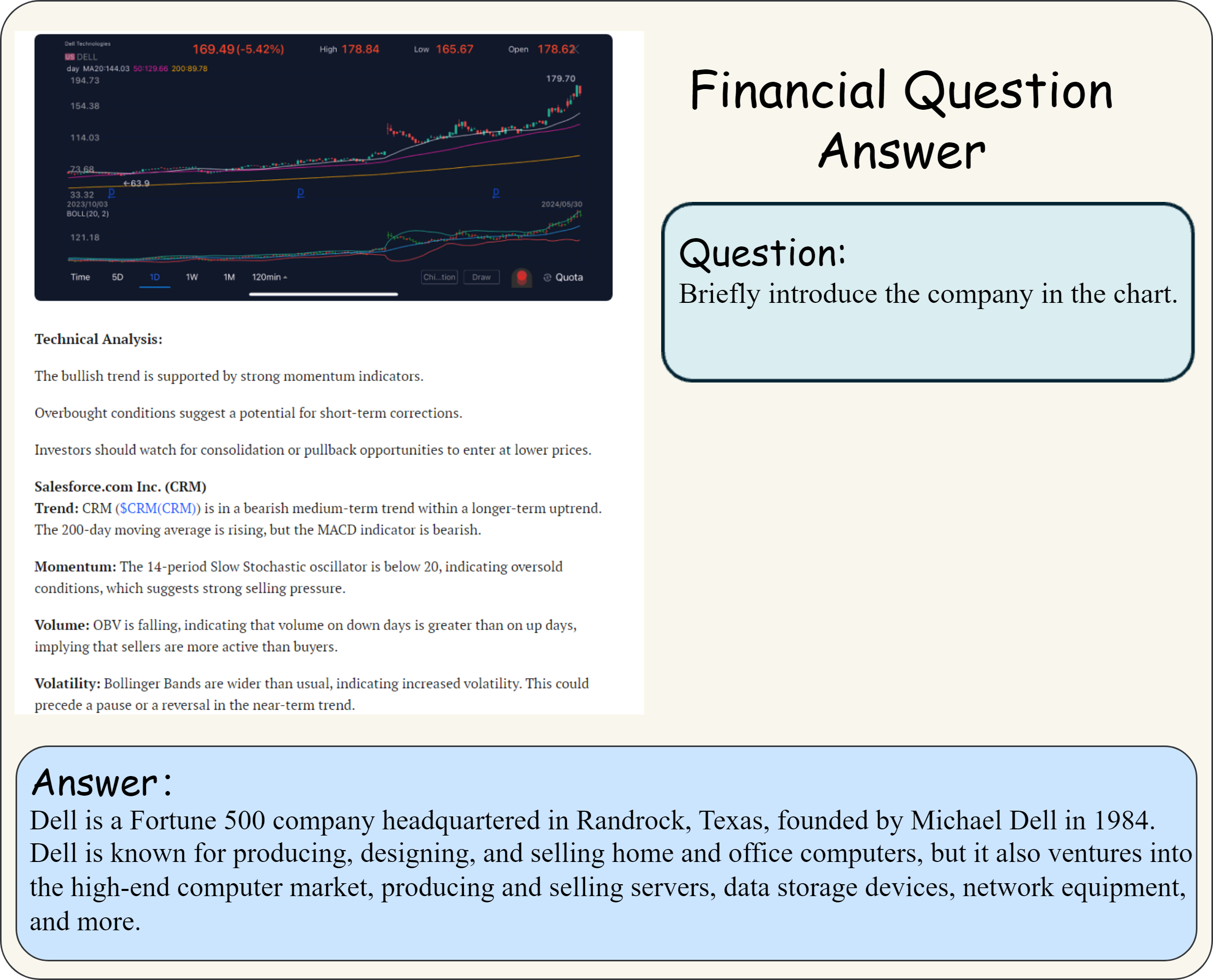}
\end{center}
\caption{Financial Question Answer.}
\label{fig: ENC}
\end{figure}

\begin{figure}[h]
\begin{center}
\includegraphics[width=.8\columnwidth]{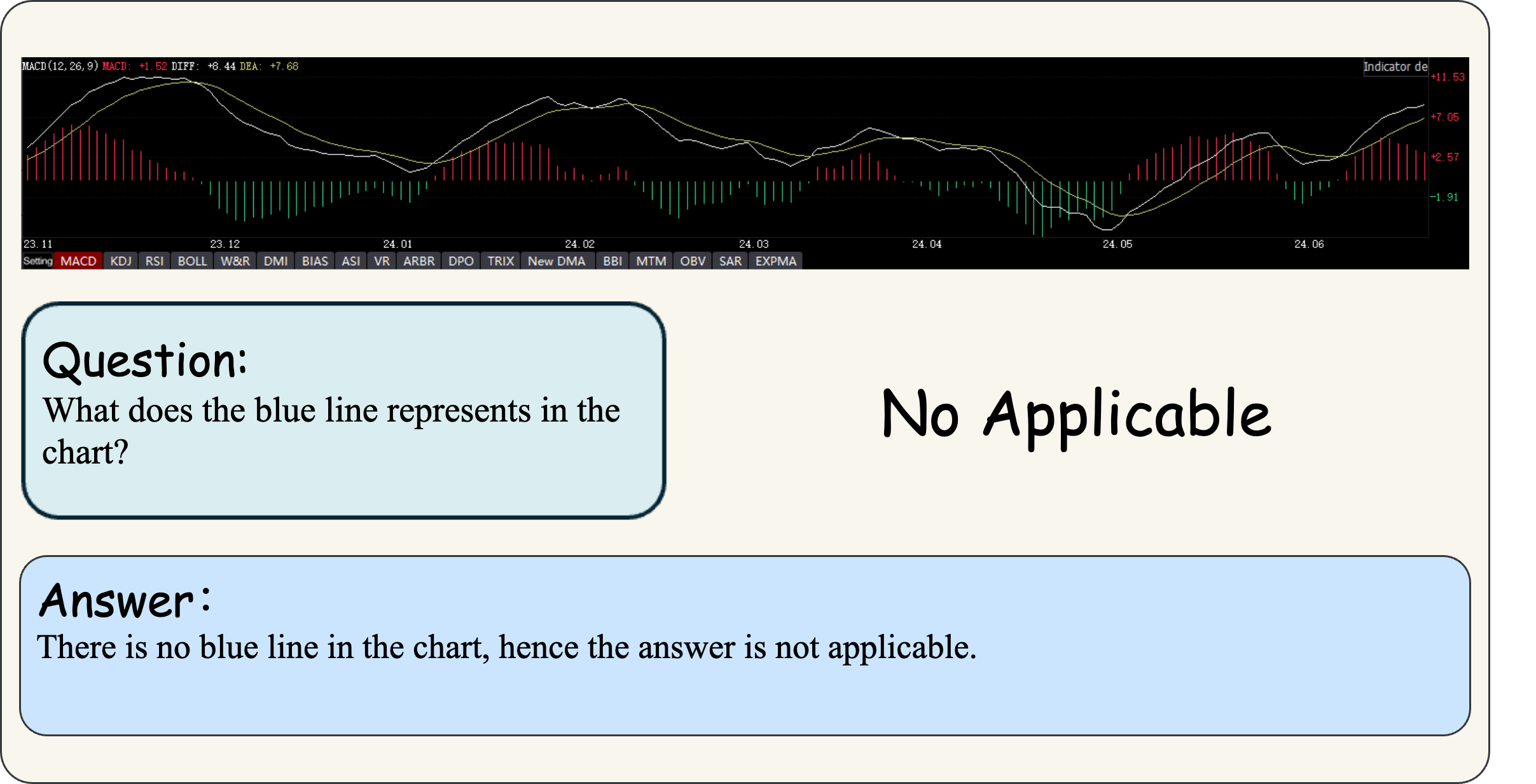}
\end{center}
\caption{Not Applicable.}
\label{fig: ENC}
\end{figure}

\end{document}